%% file: main.tex
\documentclass{article} 
\usepackage{iclr2027_conference,times}

\input{math_commands.tex}

\usepackage{hyperref}
\usepackage{url}

\usepackage{graphicx}
\usepackage{amsmath,amssymb}
\usepackage{booktabs}
\usepackage{array}
\usepackage{multirow}
\usepackage{makecell}
\usepackage{tabularx}
\usepackage{caption}
\usepackage{longtable}
\usepackage[table]{xcolor}
\usepackage{colortbl}
\usepackage{enumitem}
\usepackage[skins,breakable]{tcolorbox}
\usepackage{tikz}
\usetikzlibrary{arrows.meta,positioning,shapes.geometric,fit,backgrounds,calc}

\definecolor{tblhead}{HTML}{E7EDF3}   
\definecolor{tblalt}{HTML}{F5F7FA}    
\definecolor{hiwin}{HTML}{DCEEDD}     
\definecolor{hihot}{HTML}{FBE7D3}     
\definecolor{hicold}{HTML}{EAF1F8}    
\definecolor{cblue}{HTML}{2A7FB8}
\definecolor{cred}{HTML}{C0392B}
\definecolor{cgreen}{HTML}{3F8E5B}
\definecolor{cslate}{HTML}{34495E}
\definecolor{cgrey}{HTML}{7A8590}
\newcommand{\hw}[1]{\cellcolor{hiwin}#1}   
\newcommand{\hh}[1]{\cellcolor{hihot}#1}   
\newcommand{\na}{\text{--}}

\definecolor{algobg}{HTML}{F7F9FB}
\newtcolorbox{algobox}[1]{%
  enhanced, breakable=false, rounded corners, arc=3.5pt,
  colback=algobg, colframe=cslate!45, boxrule=0.5pt,
  left=6pt, right=6pt, top=4pt, bottom=5pt, boxsep=1pt,
  fonttitle=\bfseries\footnotesize, coltitle=cslate,
  colbacktitle=algobg, titlerule=0pt,
  attach boxed title to top left={xshift=8pt, yshift=-2.2mm},
  boxed title style={rounded corners, arc=2pt, boxrule=0.5pt,
    colframe=cslate!45, colback=white, left=4pt, right=4pt, top=1pt, bottom=1pt},
  title={#1}}
\newcommand{\kw}[1]{\textcolor{cslate}{\textbf{\footnotesize\textsf{#1}}}}

\title{Opening LLM Judges: Recovering Preference\\ Signals Beyond the Final Verdict}

\author{Sourabrata Mukherjee, Sunayana Sitaram \\
  Microsoft Research \\
  \texttt{t-somukherje@microsoft.com}}

\iclrfinalcopy 
\begin{document}

\maketitle
\lhead{}

\input{0-abstract}
\input{1-introduction}
\input{2-related}
\input{4-setup}
\input{5-gap}
\input{5-anatomy}
\input{6-evaluator}
\input{6b-consequential}
\input{7-conclusion}
\input{8-limitations}

\input{9-statements}

\bibliography{custom}
\bibliographystyle{iclr2027_conference}

\appendix
\clearpage
\input{appendix}

\end{document}

%% file: math_commands.tex
\usepackage{amsmath,amsfonts,bm}

\def\eqref#1{equation~\ref{#1}}

\def\1{\bm{1}}

\DeclareMathAlphabet{\mathsfit}{\encodingdefault}{\sfdefault}{m}{sl}
\SetMathAlphabet{\mathsfit}{bold}{\encodingdefault}{\sfdefault}{bx}{n}



%% file: 0-abstract.tex
\begin{abstract}
LLM judges are widely used to evaluate model outputs, but their verdicts can be unreliable: a judge may favor the worse answer because of its position, length, or other surface features. When a judge gives the wrong verdict, is the information needed to make the right
judgment absent from the model, or is it still present in its internal representations but not
reflected in the final output? We study this question across \textbf{64} open-weight evaluators
and \textbf{14} datasets, including causal interventions on \textbf{41} judges (editing a model's
internal activations while it runs to see whether its verdict changes). On LLMBar, a benchmark
built so that the superficially better answer is the worse one, the verdicts of $50$ judges agree
with human labels only $0.456$ of the time, even after position bias is cancelled by scoring both
answer orders. Yet a small probe trained on the same judges' internal activations, without
changing the judges, reaches $0.846$, and still $0.686$ after the influence of surface features
such as length and position is removed ($0.507$ with shuffled labels). This gap holds across
eight benchmarks and across model families, but it is not universal. A simple score of how well
surface features alone predict the human label, computed before any probe is trained, is strongly
correlated with the size of the gain (Spearman $\rho\!=\!0.90$). On two further tasks where a
judge scores one answer at a time against a rubric, leaving no surface cue to exploit, reading the
internals gives no advantage. The interventions also show that editing activations in the middle
of the network already changes the verdict, before it can be read off directly, and locate
the pathways that carry position and length bias. With the same human labels, the recovered
signal also lets a judge flag cases where it is likely to be wrong and yields better labels for
preference learning. A wrong verdict, then, does not mean the judge lacks the information, and a
simple diagnostic shows when it is worth recovering.
\end{abstract}

%% file: 1-introduction.tex
\section{Introduction}
\label{sec:intro}

LLM judges are increasingly used to evaluate model outputs, compare systems, rank models, and
generate preference data for training \citep{zheng2023judging,liu2023geval,lambert2025rewardbench};
in each case the judge exposes one small output, a preferred answer or a score, which we treat
as its decision.

That output can be unreliable. Judges respond to properties that should not determine quality,
including answer position \citep{wang2024fair}, response length
\citep{dubois2024length,saito2023verbosity}, and style \citep{panickssery2024self,sharma2024sycophancy}.
On benchmarks built so that the longer or more polished answer is the worse one, a judge can score
\emph{below chance} \citep{zeng2024llmbar}. But a model does not decide in a single token, and the
verdict is the endpoint of many layers of computation, which raises a simple question:

\begin{quote}
\emph{When an LLM judge gives the wrong verdict, is the information needed to make the right
judgment absent from the model, or is it still present in its internal representations but not
reflected in the final output?}
\end{quote}

\input{fig-flow}

We call the difference between what a judge represents internally and what its verdict exposes
the \textbf{internal--output gap}, and it can be large. On LLMBar \citep{zeng2024llmbar}, one
such benchmark, the verdicts of \textbf{50} open-weight judges agree with human labels only
$0.456$ of the time, below chance, even after position bias is cancelled by scoring both answer
orders. A small classifier, a \emph{probe}, trained on the same judges' frozen internal
activations reaches $0.846$ without changing the judges themselves. One might suspect it has
simply learned that the shorter answer wins on this benchmark, so we remove the influence of
length and eight other surface features before training it; it still reaches $0.686$, where
shuffled labels give $0.507$. The same pattern holds on eight pairwise-preference benchmarks and
is not driven by a few models or by a single model family (\S\ref{sec:gap}).

Where does the gap come from? We intervene directly on the activations of \textbf{41} judges
during a forward pass. Editing activations in the middle of the network already changes the final
verdict, earlier than the layer at which the verdict can be read off directly. Position and
length biases run through internal pathways that we can locate and switch off, and a common
attribution method points to different components than the ones whose removal actually changes
the verdict (\S\ref{sec:anatomy}). The final verdict is therefore an incomplete picture of the
computation behind it.

The gap is not universal, so we ask when reading internals is worth it. We measure how well one
surface feature, response length, predicts the human label on a benchmark, and turn the answer
into a single number, the \textbf{surface-confound score} $\mathrm{cf}$. Across the eight
benchmarks the gain from reading internals rises with $\mathrm{cf}$ (Spearman $\rho\!=\!0.90$);
this holds after accounting for judge quality and for judges that share a model family, and
$\mathrm{cf}$ from one run predicts the gain on a second, independent run. The reverse case holds
too: on two held-out benchmarks where the judge scores one answer at a time
against a rubric, so that position and length cannot tip a comparison, reading internals gives no
advantage (\S\ref{sec:apriori}). The practical rule is simple: when $\mathrm{cf}$ says the output
is likely to be unreliable, read the judge's frozen activations instead of retraining it.

Prior work has shown that relevant information can be read from an evaluator's internal
representations \citep{maiya2025preference,lai2025lager,girrbach2025latent,li2026inspector} and
that the internal components a judge uses to score can be found with causal interventions
\citep{feldhus2026judgecircuits}. Our work asks: when the emitted verdict is wrong, is the
human-preference signal still recoverable, which internal pathways keep the output from exposing
it, and can we predict when recovering it will help? We make three contributions.

\begin{enumerate}[leftmargin=1.25em,itemsep=1.5pt,topsep=2pt,parsep=0pt]
\item \textbf{A robust internal--output gap.} Across $50$ judges and eight benchmarks,
preference information recovered from frozen activations consistently improves on the emitted
verdict, including after surface controls and clustering by model family (\S\ref{sec:gap}).
\item \textbf{A causal view of evaluator decisions.} Across $41$ judges, interventions on the
forward pass locate the computation that sets the verdict and show that surface biases run
through internal pathways that attribution methods do not reliably identify (\S\ref{sec:anatomy}).
\item \textbf{A conditional and practical read-out.} A surface-confound diagnostic indicates both
where internal read-out helps and where it does not, and with the same number of human labels the
recovered signal improves evaluation and downstream preference learning
(\S\ref{sec:apriori}--\S\ref{sec:consequential}).
\end{enumerate}

Three claims must be kept apart: that preference information can be \emph{decoded} from internal
representations, that internal components \emph{causally influence} the verdict, and that the
judge actually \emph{used} that information when it decided.

%% file: fig-flow.tex
\begin{figure}[t]
\centering
\resizebox{\textwidth}{!}{%
\hyphenpenalty=10000\exhyphenpenalty=10000\relax
\begin{tikzpicture}[
  font=\scriptsize,
  box/.style={rounded corners=2.5pt, draw=cgrey!90, line width=0.5pt, align=center,
              inner sep=3.4pt},
  qbox/.style={rounded corners=2.5pt, draw=cgrey!70, fill=hicold!85, line width=0.5pt,
               align=center, inner sep=3.2pt, anchor=north west, text width=3.00cm,
               minimum height=1.26cm},
  arr/.style={-{Latex[length=1.5mm]}, draw=cslate!80, line width=0.6pt},
]
\node[box, anchor=west, text width=1.55cm, fill=tblhead] (in) at (0,-0.72)
  {\textbf{input}\\[1pt] question $q$,\\ answers $A,B$};

\node[box, anchor=north west, text width=4.05cm, align=left, draw=cslate!70, fill=cblue!5] (judge) at (2.40,0)
  {\textbf{LLM judge:} internal computation\\[1.5pt]
   \textcolor{cgreen!85!black}{$\bullet$\; preference information}\\[1pt]
   $\bullet$\; surface biases (position, length)\\[1pt]
   $\bullet$\; decision information};
\draw[arr] (in.east) -- (in.east -| judge.west);

\node[box, text width=1.02cm, anchor=west, draw=cslate!60, fill=hihot!75] (stage) at (7.18,-0.45)
  {\textbf{output stage}\\[1pt] {\tiny exposes only a verdict}};

\node[box, anchor=west, text width=3.52cm, draw=cred!75, fill=cred!6] (v) at (9.38,-0.45)
  {\textbf{final verdict} ``A''\,/\,``B''\\[1pt]
   order-averaged: \textcolor{cred}{$.456$} vs.\ humans};
\node[box, anchor=west, text width=3.52cm, draw=cgreen!75, fill=cgreen!7] (p) at (9.38,-1.72)
  {\textbf{internal read-out} \textcolor{cgreen!85!black}{$.846$}\\[1pt]
   surface features removed: $.686$};

\draw[arr] ($(judge.north east)+(0,-0.45)$) -- (stage.west);
\draw[arr, draw=cred!65!black] (stage.east) -- (v.west);
\draw[arr, rounded corners=3pt, draw=cgreen!55!black] ($(judge.south west)!0.80!(judge.south east)$)
  |- (p.west);
\node[font=\tiny, text=cgreen!40!cslate, anchor=north] at (7.20,-1.86)
  {a small probe reads the same frozen activations};

\draw[{Latex[length=1.2mm]}-{Latex[length=1.2mm]}, densely dashed, draw=cslate!85,
      line width=0.55pt] (9.70,-0.86) -- (9.70,-1.31);
\node[font=\tiny\itshape, text=cslate, anchor=west] at (9.84,-1.085)
  {the internal--output gap};

\node[qbox] (q1) at (0,-2.42)
  {\textbf{Where does the decision form?}\\[1pt] causal influence from ${\sim}.58$ of depth};
\node[qbox] (q2) at (3.4167,-2.42)
  {\textbf{Why does the verdict fail?}\\[1pt] position/length pathways; $33$--$54\%$ removable};
\node[qbox] (q3) at (6.8333,-2.42)
  {\textbf{What remains inside?}\\[1pt] preference read-out reaches $.846$};
\node[qbox] (q4) at (10.25,-2.42)
  {\textbf{When is it useful?}\\[1pt] $\mathrm{cf}$ predicts the gain and its absence, $\rho\!=\!.90$};
\foreach \n/\q in {1/q1,2/q2,3/q3,4/q4}{%
  \node[circle, fill=cslate, text=white, inner sep=1.4pt, font=\tiny\bfseries]
    at ($(\q.north west)+(0.02,0.02)$) {\n};}
\end{tikzpicture}%
}
\caption{A judge's rich internal computation reaches the user through an
output stage that exposes only a verdict. On adversarial LLMBar the verdict agrees with humans
below chance ($.456$, mean over $50$ judges), while a small probe reading the same frozen
activations reaches $.846$. The four questions below organize the paper.}
\label{fig:flow}
\end{figure}
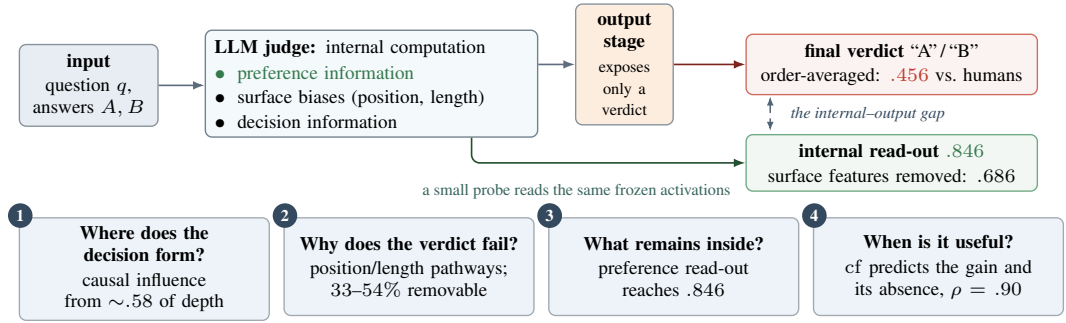

%% file: 2-related.tex
\section{Related Work}
\label{sec:related}

\paragraph{Evaluator reliability.}
LLM judges \citep{zheng2023judging,liu2023geval} and reward models
\citep{christiano2017deep,ouyang2022training} are studied behaviorally
\citep{lambert2025rewardbench,tan2025judgebench,gureja2025mrewardbench} and repaired at the
\emph{output}: preference optimization \citep{rafailov2023dpo}, multi-objective heads
\citep{wang2024armorm}, swap-averaging, self-consistency \citep{wang2023selfconsistency},
calibration \citep{guo2017calibration}. Position \citep{wang2024fair}, length
\citep{dubois2024length,saito2023verbosity} and self-preference \citep{panickssery2024self}
biases are documented, LLMBar \citep{zeng2024llmbar} shows below-chance verdicts when surface form
is decoupled from quality, and judges that agree closely with one another can still fall well short
of human agreement \citep{mukherjee2026geometry}. This line of work fixes the judge's output; it does not ask what
the judge computed internally.

\paragraph{Internal representations for evaluation.}
Recent work looks into the internal representations of the evaluator. \citet{maiya2025preference} show that
supervised and unsupervised linear probes over LLM activations extract preferences more
accurately than generation-based judgment across four model families and six datasets, and can
match fine-tuned evaluators at equal data. \citet{lai2025lager} (LAGER) aggregate score-token
logits across layers, backbone frozen, to improve \emph{point-wise} judge--human correlation.
\citet{girrbach2025latent} derive latent ratings from probability-weighted scores, verifier
probabilities, and probes at the rating position, while \citet{li2026inspector} frame evaluation
directly as representation probing in small models. Two open questions that our work addresses are \emph{why} the output loses the signal, and \emph{when}
recovering it is worthwhile.

\paragraph{Mechanistic interpretability.}
To answer the first of those we use lenses
\citep{nostalgebraist2020logitlens,belrose2023tunedlens}, activation and path patching
\citep{meng2022rome,goldowskydill2023localizing}, causal mediation \citep{vig2020causal}, causal
scrubbing \citep{chan2022causalscrubbing}, probing \citep{alain2017probing,belinkov2022probing},
and sparse features \citep{bricken2023monosemanticity,huben2024sparse,lieberum2024gemmascope},
inheriting the warning that attribution surrogates diverge from causal effects
\citep{zhang2024patching}. Concurrently, \citet{feldhus2026judgecircuits} apply edge attribution patching to judges and find
a sparse ``latent evaluator'' subgraph in mid-to-late MLPs feeding fragile format-specific output
branches. The accounts are complementary: theirs explains why one shared internal evaluator yields
different scores under different output \emph{formats}, ours asks whether the human-labeled
\emph{preference} survives the output stage at all, and when it can be recovered.

\begin{table}[t]
\centering\footnotesize
\setlength{\tabcolsep}{2.6pt}\renewcommand{\arraystretch}{1.04}
\begin{tabular}{@{}lccccccc@{}}
\toprule
\rowcolor{tblhead}
\textbf{Work} & \textbf{Internal} & \textbf{Causal} & \textbf{Surface} & \textbf{Predicts} & \textbf{Downstream} & \textbf{Judge} & \textbf{Scale} \\
\rowcolor{tblhead}
 & \textbf{read-out} & \textbf{localization} & \textbf{controls} & \textbf{when useful} & \textbf{use} & \textbf{format} & \textbf{(models)} \\
\midrule
\citet{maiya2025preference}   & \checkmark & \na & pos.\ only & \na & \na & both & $4$ families \\
\rowcolor{tblalt}\citet{lai2025lager} & \checkmark & \na & \na & \na & data sel. & point-wise & $6$ \\
\citet{girrbach2025latent}    & \checkmark & \na & \na & \na & BoN, routing & point-wise & $5$ \\
\rowcolor{tblalt}\citet{li2026inspector}      & \checkmark & \na & \na & \na & data sel. & point-wise & $4$ \\
\citet{feldhus2026judgecircuits} & (\checkmark) & \checkmark & \na & \na & \na & point-wise & $3$ families \\
\midrule
\rowcolor{hiwin}\textbf{Our work} & \checkmark & \checkmark & \checkmark & \checkmark & \makecell{\textbf{DPO labels,}\\\textbf{abstention}} & both & \textbf{64} \\
\bottomrule
\end{tabular}
\caption{Positioning of our work against the four closest lines of prior work and one
concurrent one. ``Causal localization'': forward-pass interventions that change the emitted output; ``surface controls'': the advantage is tested after removing length and position cues; ``predicts when useful'': a diagnostic computable
\emph{before} the read-out is fitted, validated on an independent run, and including a negative
case; ``downstream use'': empirically demonstrated applications (BoN: best-of-$N$); only our work
uses the recovered preferences as training labels (DPO) and to abstain on likely errors. Judge format: pairwise compares two answers in one prompt, point-wise scores one at a
time. A dash marks an axis outside that paper's scope;
(\checkmark) marks an appendix-level result. Extended discussion: App.~\ref{app:related}.}
\label{tab:related}
\end{table}

Table~\ref{tab:related} places our work against that literature along the axes taken up in turn
by \S\ref{sec:gap} (read-out under surface controls), \S\ref{sec:anatomy} (causal localization),
\S\ref{sec:apriori} (the diagnostic) and \S\ref{sec:consequential} (downstream use).

%% file: 4-setup.tex
\section{Setup}
\label{sec:setup}

A generative judge $J$ maps $(q,a_A,a_B)$ to a verdict distribution; a reward model maps
$(q,a)$ to $r\in\mathbb{R}$. Every dataset supplies a \emph{human-labeled} preference $y$ and
``accuracy'' means agreement with $y$; we treat $y$ as a noisy, annotator-derived target and use inter-annotator agreement only to
estimate its noise. We
read a judge two ways. The \textbf{verdict read-out}, what a user consumes, is the final-layer
logits over the verdict vocabulary. The \textbf{internal read-out} applies, at each layer
$\ell$, the model's own final LayerNorm and unembedding (logit lens) or a learned affine (tuned
lens) to the residual stream $h_\ell$ at the decision position, yielding a per-layer estimate
$\hat v_\ell$ and a vector for probing; \emph{lens decoding depth} is the fractional layer at
which $\hat v_\ell$ first matches the final verdict.

Because correlation is insufficient, we also intervene inside a live forward pass.
\textbf{Steering} adds $\alpha\hat u$, \textbf{ablation} projects out a subspace, and
\textbf{patching} copies activations from a counterfactual run, and each time we measure the
change in the \emph{emitted} verdict. This gives the second depth quantity we use,
\textbf{causal decision depth}: the earliest normalized layer at which a residual patch reliably
flips that verdict. It marks where our intervention first finds causally sufficient decision information; the
decision itself may still form gradually over several layers (App.~\ref{app:defs}).

These two kinds of evidence answer different questions, and we keep three claims separate
throughout. \textbf{L1 decodability} asks whether $y$ is predictable from $h_\ell$ and is
answered by probes (\S\ref{sec:gap}). \textbf{L2 causal influence} asks whether intervening on
$h_\ell$ changes the emitted verdict and is answered by patching, ablation and steering
(\S\ref{sec:anatomy}). \textbf{L3 causal use} asks whether $J$ used that direction to form
\emph{its} verdict. We establish L1 and L2 but not L3, and no claim here requires L3, since
\S\ref{sec:consequential} shows that acting on an L1 signal improves outcomes whether or not the
judge used it. Consistently, per-judge causal decision depth (L2) does not
predict per-judge probe gain (L1): $\rho\!=\!0.06$, $p\!=\!0.73$, $n\!=\!39$.

Against that background we compare five readers of the same frozen judge, each given the
\emph{same number of human labels}: (a)~the raw verdict, (b)~the order-averaged verdict, (c)~a
cross-validated probe on the internal read-out, (d)~LoRA and full-parameter fine-tunes, and
(e)~output-level fixes (self-consistency, calibrated prompting). ``Label budget'' counts only
human-labeled items; the arms differ widely in compute and storage, accounted separately in
App.~\ref{app:cost}.

The registry holds \textbf{64} open-weight models spanning $0.36$B--405B and $21$ families:
\textbf{50} generative judges forming the flagship adversarial panel, \textbf{9} scalar reward
models, and \textbf{5} read-out-only large judges (Table~\ref{tab:roster}). The causal suite
intervenes on the \textbf{41} of the 50 judges exposing hookable decoder blocks within our memory
budget, plus \textbf{5} reward models. The \textbf{14 datasets}, all previously published, divide
into \textbf{8} public pairwise-preference benchmarks carrying the ladder of \S\ref{sec:gap} and
\textbf{6} auxiliary sets supporting the mechanism and rubric analyses
(Tables~\ref{tab:datasets},~\ref{tab:aux}). Cached surrogates (lenses, direct logit attribution,
patching, causal scrubbing, SAEs, cross-validated $L_2$ probes) are validated against the
forward-pass suite. Controls include bootstrap CIs at both the judge and the
\textbf{model-family} level, paired Wilcoxon under FDR-BH, shuffled-label and surface
residualization, seed replication, and random-component null checks. Full inventory:
Table~\ref{tab:instr}; protocol: App.~\ref{app:repro}.

%% file: 5-gap.tex
\section{The Internal--Output Gap}
\label{sec:gap}

\begin{figure}[t]
\centering
\includegraphics[width=\textwidth]{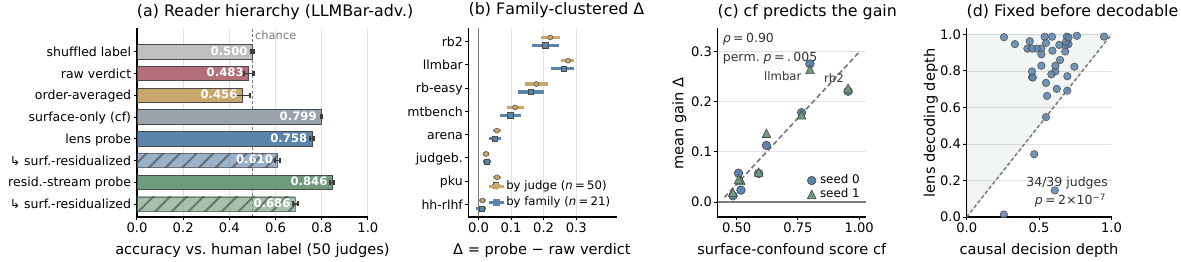}
\caption{The internal--output gap and the statistics that have to hold for it.
\textbf{(a)}~Reader hierarchy on LLMBar-adversarial ($50$ judges, bootstrap $95\%$ CIs): the
raw verdict is below chance, the surface-only baseline is high \emph{by construction} here
(\S\ref{sec:apriori}), and the internal read-outs survive surface residualization (hatched) well
above the shuffled-label control. \textbf{(b)}~Per-benchmark advantage bootstrapped over judges
and over model families; clustering widens the intervals but leaves $7/8$ excluding zero, the
exception being hh-rlhf, the lowest-cf benchmark, where \S\ref{sec:apriori} predicts the effect
should vanish. \textbf{(c)}~The diagnostic orders the gain; triangles are an independent
replication seed. \textbf{(d)}~Per judge, a residual patch alters the emitted verdict \emph{before}
a lens can decode it; causal decision depth is the onset of intervention sensitivity defined in \S\ref{sec:setup}.}
\label{fig:robust}
\end{figure}

\looseness=-1
We begin where the output stage fails most visibly. On \textbf{LLMBar-adversarial}, where the
surface-preferable answer is deliberately the worse one, the verdicts of $50$ judges, each scored on both
answer orders and averaged, reach \textbf{0.456} accuracy against the human label: below chance, below the raw verdict's
$0.483$ (order-averaging strips a position bias that was partly masking the length bias, so it
\emph{lowers} accuracy here), and far below a prefer-shorter heuristic ($0.799$) we confront in
\S\ref{sec:apriori}. $72\%$ of judges are below chance although the same judges reach a median
$0.89$ on the easy split, and averaging across all layers does no better ($0.461$). Yet a within-adversarial cross-validated \textbf{probe on the mid-layer
read-out reaches $0.758$}, positive for $100\%$ of judges, and reading the residual stream reaches
\textbf{0.846} (Fig.~\ref{fig:robust}a). Seed $1$ reproduces each of these within $\pm0.012$
(cell-level $r\!=\!0.97$ over all $400$ cells).

The obvious objection is that the probe has simply learned answer length, which predicts the label
on this set. Three controls separate content from surface. \textbf{(i)~Residualization:} regressing
out a nine-dimensional surface feature vector (length, position, formatting and lexical markers)
before fitting leaves the lens probe at $0.610$ and the residual-stream probe at
$\mathbf{0.686}$, against shuffled-label controls of $0.500$/$0.507$, and the residualized
read-out beats its shuffled control for $50/50$ judges ($p\!<\!10^{-15}$).
\textbf{(ii)~Length balancing:} on a length-balanced subset the probe scores $0.684$ where the
order-averaged verdict scores $0.482$ ($p\!=\!4{\times}10^{-10}$). \textbf{(iii)~Capacity:} the
residual-stream probe, which strictly contains the lens probe's features, beats it on $7/8$
benchmarks, so the lens number understates what the activations contain. Human-label-predictive signal therefore survives
the \emph{measured} surface controls. We call what remains the \emph{surface-controlled preference signal}. It establishes decodability (L1)
only, and it controls for the surface features we measured.

\label{sec:ladder}
The pattern holds well beyond one adversarial set. Across \emph{arena, rb-easy, mtbench, hh-rlhf,
pku-safe, judgebench, rewardbench2, llmbar-adv} ($50$ judges each; $400$ cells), the probe beats
the raw verdict on \textbf{all 8} benchmarks (mean $+0.117$; Table~\ref{tab:ladder}a), improving
$89\%$ of individual cells.\footnote{On MT-Bench the residual-stream probe reaches $0.756$, above
that set's $0.646$ human--human agreement. Inter-annotator agreement estimates label noise; it
does not bound accuracy attainable against one annotator-derived label set.} The distribution
matters more than the mean: the median cell gain is $+0.075$ (IQR $[+0.023,+0.209]$,
$p_{90}\!=\!+0.298$) and all $50$ judges gain on average (median $+0.119$), so no small set of
pathological judges carries the effect.

A stricter worry is independence, since the $50$ judges come from only $\mathbf{21}$ model families. Resampling by \emph{family} (Fig.~\ref{fig:robust}b) shrinks the
pooled estimate only slightly, $+0.117$ $[0.105,0.129]$ to $+0.108$ $[0.089,0.127]$, and leaves it
positive in \textbf{all $21$ families} ($150/168$ family$\times$benchmark cells); family-clustered
CIs exclude zero on $7$ of $8$ benchmarks and all $8$ survive a family-level paired Wilcoxon under
FDR-BH. The one interval covering zero (hh-rlhf, $+0.010$ $[-0.008,+0.024]$) is the
lowest-confound benchmark in the ladder, exactly what \S\ref{sec:apriori} predicts.

\begin{table}[t]
\centering
\begin{minipage}[t]{0.494\textwidth}
\centering\footnotesize
\textbf{(a) Evaluator ladder} (50 judges $\times$ 8 benchmarks)\\[2pt]
\setlength{\tabcolsep}{1.2pt}\renewcommand{\arraystretch}{1.23}
\begin{tabular}{@{}l@{\hspace{5pt}}ccccccc@{}}
\toprule
\rowcolor{tblhead}
\textbf{Bench.} & \textbf{cf} & \textbf{raw} & \textbf{o\text{-}avg} & \textbf{probe} & \textbf{resid} & \textbf{$\Delta_{\text{r}}$} & \textbf{$\delta_{\text{o}}$} \\
\midrule
\rowcolor{hihot}rb2 & .96 & .620 & .684 & .840 & \hw{\textbf{.907}} & $+.22$ & \hw{$+.16$} \\
\rowcolor{hihot}llmbar & .80 & .483 & .456 & .758 & \hw{\textbf{.846}} & $+.28$ & \hw{$+.30$} \\
rb-easy      & .77 & .690 & .815 & .868 & \hw{\textbf{.879}} & $+.18$ & $+.05$ \\
\rowcolor{tblalt}mtbench      & .62 & .626 & .697 & .739 & \hw{\textbf{.756}} & $+.11$ & $+.04$ \\
arena        & .59 & .566 & .616 & .623 & \hw{\textbf{.631}} & $+.06$ & $+.01$ \\
\rowcolor{tblalt}judgebench   & .52 & .516 & .545 & .540 & \hw{\textbf{.548}} & $+.02$ & $-.01$ \\
pku-safe     & .51 & .564 & .608 & .622 & \hw{\textbf{.647}} & $+.06$ & $+.01$ \\
\rowcolor{tblalt}hh-rlhf      & .49 & .551 & \hw{\textbf{.582}} & .563 & .555 & $+.01$ & $-.02$ \\
\midrule
mean, seed 0 & .66 & .577 & .625 & .694 & \hw{\textbf{.721}} & $+.12$ & $+.07$ \\
\rowcolor{tblalt}mean, seed 1 & .65 & .576 & .626 & .697 & \hw{\textbf{.727}} & $+.12$ & $+.07$ \\
\midrule
\rowcolor{tblhead}\textbf{beats raw} & & ref. & 7/8 & \textbf{8/8} & \textbf{8/8} & & \\
\rowcolor{tblhead}\textbf{fam.\ CI$>$0} & & ref. & 7/8 & \textbf{7/8} & \textbf{7/8} & & \\
\rowcolor{tblhead}\textbf{beats o-avg} & & \na & ref. & 6/8 & \textbf{7/8} & & \\
\bottomrule
\end{tabular}
\end{minipage}\hfill%
\begin{minipage}[t]{0.494\textwidth}
\centering\footnotesize
\textbf{(b) Forward-pass causal suite} (41 judges)\\[2pt]
\setlength{\tabcolsep}{1.2pt}\renewcommand{\arraystretch}{1.23}
\begin{tabular}{@{}lc@{}}
\toprule
\rowcolor{tblhead}\textbf{Quantity} & \textbf{Value} \\
\midrule
\rowcolor{hicold}\multicolumn{2}{@{}l}{\emph{Where the decision is established}} \\
\quad Causal decision depth (frac.) & \hh{$.58$} \\
\quad Lens decoding depth ($34/39$ later) & \hw{$.91$} \\
\quad Steering slope (20/50/90\%) & $.13/.47/1.70$ \\
\rowcolor{hicold}\multicolumn{2}{@{}l}{\emph{Attribution $\neq$ causation}} \\
\quad Attribution vs.\ cause $\rho$ & \hh{$.31$} \\
\quad Indirect (mediated) share & \hh{$82\%$} \\
\quad\quad MLP- vs.\ attention-mediated & $.63/.39$ \\
\quad Surrogate vs.\ direct $\rho$ ($40/41$) & \hw{$.78$} \\
\rowcolor{hicold}\multicolumn{2}{@{}l}{\emph{Bias is a localizable circuit}} \\
\quad Position heads (random ctrl.) & $33\text{--}54\%\ (.03)$ \\
\quad Length subspace ablation & $.41\!\to\!-.11$ \\
\quad Debias transfer (unseen data) & $-51\%$ \\
\quad Steering flips emitted verdict & $67\%$ \\
\bottomrule
\end{tabular}
\end{minipage}
\caption{(a)~The evaluator ladder: mean accuracy against human labels over $50$ judges, ordered
by the surface-confound diagnostic \textbf{cf}, with $\Delta_{\text{r}}\!=\!$~probe$-$raw and
$\delta_{\text{o}}\!=\!$~probe$-$order-average, the probe's non-redundant value; the footer gives the
eight-benchmark mean on seed $0$ and on the independent replication seed $1$, and the number of
benchmarks whose family-clustered CI for the gain over raw excludes zero. (b)~Headline
quantities of the forward-pass causal suite of \S\ref{sec:anatomy}, seed-stable across seeds
$0/1/2$ and measured inside a live forward pass. Amber marks the high-confound regime and the
three central mechanistic numbers; green marks the best reader per row, the per-judge depth
ordering, and the surrogate validation. cf is a diagnostic of
\emph{where} to read internals (\S\ref{sec:apriori}).}
\label{tab:ladder}
\end{table}

%% file: 5-anatomy.tex
\section{The Anatomy of a Judge}
\label{sec:anatomy}

At which layer do a judge's activations start to determine its verdict?
Unless noted, the causal numbers below come from the forward-pass suite on \textbf{41
judges} ($3000{+}$ measurements), are seed-stable across seeds $0/1/2$ (Table~\ref{tab:ladder}b),
and agree with cached surrogates for $40/41$ judges ($\rho\!=\!0.78$).

Editing activations changes the emitted verdict at earlier layers than those at which the verdict
can first be read off. A residual patch first flips the verdict at a median of $0.58$ of
normalized depth, and steering moves it from the middle layers onward, with an effect that grows
with depth (Table~\ref{tab:ladder}b).\footnote{Causal onset is estimated on a coarse $10$-point
depth grid from a small sample of paired items per judge (App.~\ref{app:repro}), so it locates the
decision only approximately.} The same order holds for individual judges: on $\mathbf{34}$ of the
$\mathbf{39}$ judges with both measurements, patching changes the verdict at an earlier layer than
the one at which a lens can decode it (median causal decision depth $0.585$ vs.\ lens $0.906$;
paired Wilcoxon $p\!=\!2{\times}10^{-7}$; Fig.~\ref{fig:robust}d). For rubric scoring of a single
answer, the decision forms later, at $0.9$--$1.0$ of depth. The two depths are consistent: a
middle layer can hold the information that decides the verdict in a form that later layers already
use but that a linear read-out cannot yet recover. On easy items the tuned lens and the logit lens
find the verdict at the same depth, so no extra fitting is needed; on adversarial items the
untrained read-out is wrong at every depth, so a trained probe is required. Causal decision depth
also does \emph{not} predict which judges gain most from the probe ($\rho\!=\!0.06$,
$p\!=\!0.73$): how early a verdict can be changed and how much human preference can be recovered
are separate properties of a judge (\S\ref{sec:setup}).

\label{sec:attrcause}
Locating the computation this way also exposes a methodological trap. Direct logit attribution
(DLA), the additive surrogate in common use, is an \emph{incomplete proxy for causal influence}
in our intervention suite. Correlating DLA with a causal-knockout ranking over $41$ judges gives
$\rho\!=\!0.31$ at the component level
(CI $[0.30,0.33]$; below $0.5$ for $95\%$ of judges; peak-attribution layer $30$ vs.\ peak-cause
layer $18$). Mediation explains why: $\mathbf{82\%}$ of the biased effect is indirect (CI
$[0.81,0.84]$), and in separate decompositions of that indirect effect MLP-mediated contributions
($0.63$) exceed attention-mediated ones ($0.39$).\footnote{The two fractions come from separate decompositions and need not sum to one, so we
only compare them: MLP-mediated effects are \emph{larger than} attention-mediated ones.} Attribution sees the direct
path and is blind to the mediated majority, and consistently the cached surrogate agrees better
with the direct ($\rho\!=\!0.78$) than the total effect ($0.66$). The same holds for scalar
reward models ($\rho\!=\!0.32$; an earlier cached $\rho\!\approx\!0.97$ there was a random-head
artifact our null check caught).

What the mediated pathway carries is, in large part, surface bias, and that bias can be located
and removed by intervention (Table~\ref{tab:ladder}b). \textbf{Position bias} is mediated in part
by a small set of attention heads: ablating the top-ranked heads reduces the bias gap by
$33$--$54\%$, against $\approx\!3\%$ for random heads, and heads selected this way on one benchmark reduce it by $51\%$ on other benchmarks. \textbf{Length bias} is linearly represented: we extract a length direction from each
judge's residual stream, and projecting out that subspace removes the bias ($0.41\!\to\!-0.11$).
Removing it alone leaves agreement with human labels unchanged, so the usable preference signal has
to be recovered by reading the representation (\S\ref{sec:gap}). Directions extracted from one
judge also shift a \emph{different} judge's score, so judges share the geometry of the bias, though
each encodes it in its own coordinates. Steering along these directions flips the emitted verdict
in $\approx\!67\%$ of cases, so the directions change the judge's output itself.

A final property gives the internal--output gap a geometric counterpart: the output exposes a
lower-dimensional projection of a richer internal state (effective rank
$\approx$$2$--$4$, aligned only weakly with the human axis), and scalar reward models behave the
same way, with a mid layer reading an RM's reward $+0.2$--$0.3$ higher in Spearman $\rho$ with the
human label than the scalar head does (App.~\ref{app:mech}).

%% file: 6-evaluator.tex
\section{When Reading Internals Helps}
\label{sec:apriori}

The gap is conditional, and making the condition measurable turns the read-out from a universal
recommendation into a falsifiable one. The deployable recipe (App.~\ref{app:recipe}) uses only
the judge's own forward pass and a small labeled set, with no weight updates: cache the residual
stream over both answer orders, order-average to cancel position bias, and fit a small
$L_2$-logistic probe on the logit-lens read-out.

We measure the severity of this failure mode with $\mathrm{cf}$, the cross-validated accuracy of a
length-only classifier on the evaluation set. It needs no internal read-out and no probe, only the small labeled sample of that distribution
that the read-out would use anyway, so we call it \emph{pre-read-out}. Ordered by $\mathrm{cf}$
(Table~\ref{tab:ladder}a), the eight benchmarks line up with the gain the probe obtains at
Spearman $\rho\!=\!0.90$, with an \textbf{exact permutation test} over all $8!\!=\!40{,}320$
orderings giving $p\!=\!0.005$ (Pearson $r\!=\!0.91$, Kendall $\tau\!=\!0.79$). Eight
benchmark-level points are only a summary, so we add three checks. The relation is not simply that $\mathrm{cf}$ marks weak judges: after partialling out mean judge accuracy the
correlation is still $\rho\!=\!0.90$ ($p\!=\!0.002$), while judge accuracy alone predicts the
gain at only $\rho\!=\!0.21$. The correlation stays between $0.86$ and $0.96$ when any one
benchmark is dropped, and the genuinely held-out test is an independent replication run:
$\mathrm{cf}$ computed on seed $0$ predicts the gains measured on seed $1$ at $\rho\!=\!0.95$
(Fig.~\ref{fig:robust}c). Finally it holds at the cell level, where the $400$ cells give far more evidence than the eight means: fitting
$\Delta_{ij}=\beta_0+\beta_1\mathrm{cf}_j+u_{\text{family}(i)}+\epsilon_{ij}$ gives
$\hat\beta_1\!=\!0.53$ with a family-clustered $95\%$ CI of $[0.46,0.59]$, unchanged under a family random-intercept mixed model ($0.53$) and Huber-robust estimation ($0.54$);
adding judge accuracy as a covariate even raises it ($0.60$).

On the two most confounded benchmarks the length-only classifier is itself more accurate than
the probe ($0.80$ vs.\ $0.758$ on LLMBar; $0.96$ vs.\ $0.907$ on RB2). This is expected: those benchmarks were built so that length predicts the label, and the
length-only classifier is our measure of that failure mode. It would also make a poor judge: it
needs the labels of the very benchmark it scores and drops to chance where length carries no
signal ($0.52$ on JudgeBench), while the probe keeps a clear advantage once length and the other
surface features are removed (\S\ref{sec:gap}). We also do not claim that $\mathrm{cf}$ predicts when internal read-out helps
in every setting; it is a diagnostic for one common failure mode, surface confounding, in the
benchmarks studied here.

\label{sec:rubricnull}
A diagnostic should also say where the effect will be absent, so we test $\mathrm{cf}$ in three
settings of increasing difficulty. \textbf{(i)~Clean benchmarks.} The probe's non-redundant value
over the free order-averaging debias ($\delta_{\text{o}}$, Table~\ref{tab:ladder}a) concentrates
on the confounded sets ($+0.30$ llmbar, $+0.16$ rb2) and falls to $\approx\!0$ elsewhere; the two
negative cells, judgebench ($-0.01$) and hh-rlhf ($-0.02$), are the two lowest-$\mathrm{cf}$
benchmarks, and hh-rlhf is also the one whose family-clustered interval covers zero.
\textbf{(ii)~Zero-label cross-judge transfer.} A read-out fitted on \emph{other} judges and
applied to an unseen judge with none of its labels improves human agreement by $+0.08$ on LLMBar
(beating that judge's verdict for $78\%$ of judges) but is flat to slightly negative on JudgeBench
($-0.01$) and RB2 ($-0.03$). A combined internal-plus-length reader reaches a much larger $+0.32$,
but that figure is largely carried by the length feature, a within-benchmark oracle there, so we
report the internals-only number and keep the combined one in App.~\ref{app:eval}. \textbf{(iii)~A held-out task format.} Absolute rubric scoring presents
one answer at a time, so no position or length cue separates candidates and our account predicts
\emph{no} internal advantage. We tested this on two existing human-labeled Indic rubric benchmarks
($50$ and $44$ judges, $n\!=\!80$ graded human scores each), a task format and pair of benchmark
sets held out from every other claim here. It holds: the read-out does no better than the emitted score (Samiksha $\rho\!=\!-0.006$ final, $-0.005$ best layer, $+0.017$ probe; Pariksha
$+0.013$, $+0.003$, $-0.053$), and a held-out best layer beats the final layer on $52\%$ and $50\%$
of judges, matching the $50\%$ predicted.

This null result needs reconciling with one prior finding. \citet{lai2025lager} report the opposite sign
for point-wise scoring, where cross-layer logit aggregation \emph{improves} judge--human
correlation. We do not read our null as a contradiction: the settings differ on four axes
(pairwise vs.\ point-wise, adversarial confounding vs.\ ordinary scoring, a trained probe vs.\
training-free aggregation, and a different read-out target), and our account predicts internal
read-out should help whenever the \emph{output stage} is the locus of error. Miscalibrated
score-token distributions are another such case, as \citet{girrbach2025latent} observe in finding
that probes recover signal specifically where output logits are miscalibrated. So our claim is not that
internal representations are always better. Reading internals helps when the judge's output stage
is where the error enters, and $\mathrm{cf}$ measures one common way that happens.

Two deployment properties follow. The reader lifts the floor: the weakest quartile of judges (mean
raw $0.50$) reaches $0.65$, narrowing the gap to the strongest quartile from $0.17$ to $0.08$,
while probe accuracy still rises with judge quality (slope $+0.56$), so weak judges are helped
without a ceiling being hit. Internal confidence also supports \emph{abstention},
gaining $+0.39$ risk--coverage AUC over the verdict margin on LLMBar, where that margin
is \emph{below chance} ($0.431$; Fig.~\ref{fig:conseq}d). The reader is judge-portable but
distribution-specific, and it reorders \emph{which} judge one would deploy (App.~\ref{app:eval}).

%% file: 6b-consequential.tex
\section{Downstream Consequences}
\label{sec:consequential}

Holding the human-label budget fixed across arms, the read-out beats output-level debiasing and
the tested full-parameter fine-tuning recipe, without the latter's safety regression, and
produces better preference labels, with each advantage scaling with the same diagnostic. The
read-out here is the $41$-parameter logit-lens probe of App.~\ref{app:recipe}.

\begin{figure}[t]
\centering
\includegraphics[width=\textwidth]{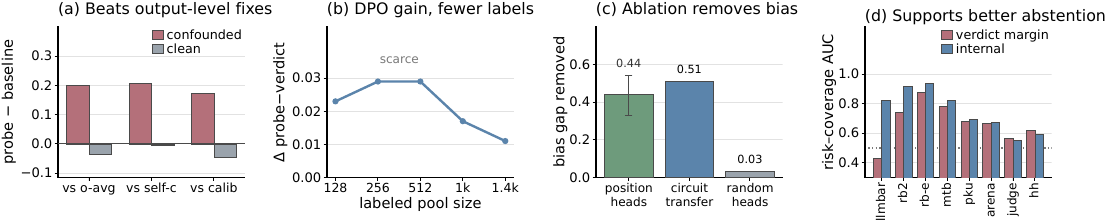}
\caption{Four consequences of reading internals, at a matched human-label budget.
\textbf{(a)}~The probe's edge over output-level fixes is confined to confounded cells and
vanishes, or reverses, on clean ones. \textbf{(b)}~Its DPO labeling advantage grows as labels
get scarce. \textbf{(c)}~Ablating the position-bias heads removes much of the bias, also on unseen data.
\textbf{(d)}~Internal confidence supports better abstention than the judge's own verdict margin,
which on LLMBar is \emph{below} chance. Full ten-panel version: Fig.~\ref{fig:conseqfull}.}
\label{fig:conseq}
\end{figure}

We first compare against output-level fixes. Giving every arm the \textbf{same 256 human labels} over
\textbf{24 cells} (6 judges $\times$ 4 benchmarks, same items), we compare the probe to
\emph{self-consistency} ($k\!=\!5$, majority-voted) and a \emph{calibrated prompt} telling the
judge to ignore length and position (Table~\ref{tab:conseq}a). The probe wins on mean accuracy,
with its advantage concentrated on the confounded cells: $+0.17$ to $+0.21$ over every
output-level fix on LLMBar and RB2, $\approx\!0$ on clean JudgeBench and Arena. Sampling more
verdicts does not close this gap.

\looseness=-1
A stronger alternative is to retrain the judge. We full-FT a 24B roster judge
(Mistral-Small-24B) on the verdict-format objective across $2\times8$ B200s (FSDP2) and compare
it to the frozen-judge probe fit on the \emph{same} $3{,}349$ human labels, scored on the same
untouched $590$-item test set (Table~\ref{tab:conseq}b). The frozen probe reaches higher overall accuracy
($0.753$) than both the fine-tune ($0.659$) and the base judge ($0.700$), with no weight updates.
Since all readers score the same items we test the probe against the base with an exact McNemar
test on the $590$ paired predictions: $+0.049$ $[+0.017,+0.083]$, $p\!=\!0.005$.\footnote{The paired test
uses cached base verdicts scoring $0.703$ (two unparsed items resolved differently from the table's
$0.700$); per-item fine-tune predictions were not retained, so the $+0.094$ gap over it has no
paired test.} The fine-tune mainly teaches the judge to
answer in the expected format (parse rate $1.0$ vs.\ $0.003$). Outside the safety split it gains
only $+0.025$, and on pku-safe its accuracy falls from $0.716$ to $0.569$, while the probe leaves
the base model unchanged and keeps its safety accuracy ($0.707$). We used a standard recipe without
tuning its hyperparameters, so a tuned fine-tune may avoid this drop, and we do \textbf{not} claim
that probing beats a well-tuned fine-tune in general. At matched labels the probe recovers about half of the mean LoRA gain
($0.550\!\to\!0.634$ vs.\ LoRA $0.724$), and most of it on confounded LLMBar
($0.437\!\to\!0.774$ vs.\ $0.881$), learning $41$ parameters where the fine-tune updates $23.6$ billion
(App.~\ref{app:cost}).

\begin{table}[t]
\centering\footnotesize
\begin{minipage}[t]{0.295\textwidth}
\centering
\textbf{(a) Output-level fixes}\\[2pt]
\setlength{\tabcolsep}{2.2pt}\renewcommand{\arraystretch}{1.10}
\begin{tabular}{@{}lcc@{}}
\toprule
\rowcolor{tblhead}
\textbf{Reader} & \textbf{mean} & \textbf{conf.} \\
\midrule
raw verdict            & 0.568 & \hh{ref.} \\
order-average          & 0.600 & \hh{$+.20$} \\
self-consist.\ $k{=}5$ & 0.581 & \hh{$+.21$} \\
calibrated prompt      & 0.617 & \hh{$+.17$} \\
\rowcolor{hiwin}internal probe & \textbf{0.682} & \emph{best} \\
\bottomrule
\end{tabular}
\end{minipage}\hfill%
\begin{minipage}[t]{0.305\textwidth}
\centering
\textbf{(b) Full fine-tuning}\\[2pt]
\setlength{\tabcolsep}{2.2pt}\renewcommand{\arraystretch}{0.95}
\begin{tabular}{@{}lccc@{}}
\toprule
\rowcolor{tblhead}
\textbf{Bench} & \textbf{base} & \textbf{FT} & \textbf{probe} \\
\midrule
rb2        & 0.787 & 0.818 & \hw{\textbf{0.893}} \\
llmbar     & 0.511 & \textbf{0.660} & \textbf{0.660} \\
judgebench & 0.548 & 0.495 & \hw{\textbf{0.570}} \\
\rowcolor{hihot}pku-safe & \textbf{0.716} & 0.569 & \hw{0.707} \\
\rowcolor{hiwin}\textbf{overall} & 0.700 & 0.659 & \textbf{0.753} \\
excl.\ pku & 0.690 & 0.715 & \textbf{0.781} \\
\bottomrule
\end{tabular}
\end{minipage}\hfill%
\begin{minipage}[t]{0.395\textwidth}
\centering
\textbf{(c) DPO preference labels}\\[2pt]
\setlength{\tabcolsep}{2.2pt}\renewcommand{\arraystretch}{1.10}
\begin{tabular}{@{}llccc@{}}
\toprule
\rowcolor{tblhead}
\textbf{pool$\cdot$judge} & \textbf{pol.} & \textbf{verd.} & \textbf{probe} & \textbf{gold} \\
\midrule
\rowcolor{hihot}llmbar$\cdot$q7b & 1.5b & 0.463 & \hw{\textbf{0.787}} & 0.825 \\
\rowcolor{hihot}llmbar$\cdot$q7b & 3b & 0.625 & \hw{\textbf{0.775}} & 0.863 \\
rb2$\cdot$q7b & 1.5b & 0.906 & \hw{0.931} & 0.960 \\
rb2$\cdot$g9b & 1.5b & 0.940 & \hw{0.951} & 0.963 \\
rb2$\cdot$q7b & 3b & 0.923 & \hw{0.934} & 0.963 \\
\bottomrule
\end{tabular}
\end{minipage}
\caption{Three matched-budget comparisons: (a)~readers over $24$ cells at $256$ labels, with the
probe's paired advantage on the \textbf{conf}ounded cells ($\approx\!0$ on clean ones);
(b)~the frozen probe against a \emph{full}-parameter fine-tune of a 24B judge, same $3{,}349$
labels and same held-out $590$ items, safety regression in amber; (c)~probe- against
verdict-labeled DPO, gold-test accuracy, over five pool$\cdot$judge$\cdot$policy cells. Per-row
$n$, per-cell $\Delta$, and the label-efficiency sweep: App.~Tables~\ref{tab:guide}
and~\ref{tab:er8}.}
\label{tab:conseq}
\end{table}

Better evaluation matters only if it changes what one can build, so we finally ask whether the
recovered signal produces better training data. We label a preference pool three ways (verdict,
probe, gold), DPO-tune a small policy on each, and evaluate on a held-out gold split
(Table~\ref{tab:conseq}c). Probe labels win on four of five cells and are level on the fifth. On
adversarial LLMBar the verdict labels are near-random ($0.53$), so verdict-DPO lands below chance
($0.463$); on that 1.5B cell \textbf{probe labels recover $\approx\!90\%$ of the verdict-to-gold
gap} ($0.463\!\to\!0.787$, gold $0.825$, non-overlapping CIs, $n\!=\!80$), and the gain survives a
$1.5\text{B}\!\to\!3\text{B}$ policy swap ($+0.15$). We do not claim probe labels equal human
labels: on less-confounded RB2 the gain is small ($+0.016$) but grows as labels get scarce.

\label{sec:coherence}
A single quantity ties the section together: ranked by $\mathrm{cf}$, the accuracy gain, the
abstention advantage, the same-label advantage and the DPO gain all decline together and vanish on
length-neutral sets (App.~Table~\ref{tab:coherence}). In our eight benchmarks, reading internals pays
off once $\mathrm{cf}\!\gtrsim\!0.75$, a rule of thumb that new settings should recheck.

%% file: 7-conclusion.tex
\section{Conclusion}
\label{sec:conclusion}

LLM judges are usually treated as black boxes whose verdict is the evaluation, and our results
show that this can leave useful information on the table. Across $64$ open-weight evaluators, a judge's
final verdict can be wrong while the information needed to agree with humans remains recoverable
from its internal activations: a small probe on those frozen activations clearly beats the judge's
own verdict on adversarial pairwise evaluation, the gain holds after controlling for surface
features and model family, and causal interventions locate the pathways behind the bias.

Reading the internals is not always better: the advantage concentrates where surface features
strongly predict the human labels and largely disappears on rubric tasks without those confounds,
so it is best used as a targeted diagnostic and repair alongside LLM judging. When a judge's output
is unreliable, useful preference information may still be inside the model, and a small probe can
recover it without changing the model.

%% file: 8-limitations.tex
\paragraph{Limitations.}
We show that preference information is recoverable (L1) and that internal components causally
influence the verdict (L2); whether the judge used it (L3) is open (App.~\ref{app:limits}).

%% file: 9-statements.tex
\clearpage

\subsection*{AI use statement}

We used general-purpose AI assistants at several points in this work, in every case under author
review and revision. In \textbf{writing}, they helped polish prose, tighten phrasing, and improve
the readability and consistency of the manuscript; the claims, framing, and scientific content
are the authors' own. In \textbf{argumentation}, we used them as a sounding board to stress-test
a few arguments and to surface counter-arguments, which we then checked against our own results.
In \textbf{implementation}, they produced initial drafts of experiment, analysis, and plotting
code, which the authors subsequently reviewed, debugged, and substantially rewrote; the released
code reflects that editing rather than any generated draft. They also assisted with
\textbf{presentation}, such as \LaTeX{} table and figure layout, formatting, and visual design.
Every number and claim in the paper was verified by the authors against the released result
artifacts. AI assistants were not used to generate any central or core experimental data or
labels. LLMs appear in this work as the \emph{research objects} under study
(\S\ref{sec:setup}, \S\ref{app:roster}), which is a separate matter from their assistive use
described here.

\subsection*{Ethics statement}

This work studies existing open-weight models on existing, previously published datasets. It
collects no new human data and involves no human subjects. Every model and dataset is used under
the terms set by its creators, and we redistribute no model weights and no underlying corpus. One
evaluation set is a restricted human-annotated corpus, so we release its schema, prompts, and the
results computed from it, but not the corpus itself (\S\ref{app:limits}). Human preference labels
carry the judgments and the biases of the people who wrote them, so we treat agreement with those
labels as a noisy target rather than as ground truth and report inter-annotator agreement where
it is available. Finally, the internal directions we use to locate and switch off a judge's
position and length bias could in principle be steered the other way to strengthen that bias. We
report them so that evaluator failures can be diagnosed and repaired, and we state that risk here
rather than leave it implicit.

\subsection*{Reproducibility statement}

The deployable read-out recipe is stated as Algorithm~1 in \S\ref{app:recipe}; the model
registry, dataset sheet, and instrument inventory are in \S\ref{app:roster}; the fine-tuning and
probe configurations, seeds, and determinism settings are in \S\ref{app:repro}; and the analytic
cost accounting behind the fine-tuning comparison is in \S\ref{app:cost}. We release the code,
the model registry, per-experiment result artifacts, the evaluation prompts, the 24-cell
\textsc{Budget} grid, and the multi-node full-parameter fine-tuning configuration; see
\S\ref{app:repro}.

%% file: appendix.tex
\section{Appendix}
\label{sec:appendix}

Table~\ref{tab:appguide} maps this appendix: what each section is for, and what each table and
figure in it shows.

{\footnotesize
\setlength{\tabcolsep}{3pt}\renewcommand{\arraystretch}{1.15}
\begin{longtable}{@{}>{\raggedright\arraybackslash}p{0.235\columnwidth}>{\raggedright\arraybackslash}p{0.305\columnwidth}>{\raggedright\arraybackslash}p{0.395\columnwidth}@{}}
\caption{Guide to the appendix: section purposes, and what each table and figure shows.}
\label{tab:appguide}\\
\toprule
\rowcolor{tblhead}\textbf{Section} & \textbf{What it is for} & \textbf{Tables and figures} \\
\midrule
\endfirsthead
\toprule
\rowcolor{tblhead}\textbf{Section} & \textbf{What it is for} & \textbf{Tables and figures} \\
\midrule
\endhead
\bottomrule
\endfoot
\S\ref{app:defs} Additional Definitions & Definitions of dimensional collapse and of the concept directions used in steering and ablation. & \na \\
\rowcolor{tblalt}\S\ref{app:claims} Claim--Evidence Matrix & Every load-bearing claim against the evidence for it. & Tab.~\ref{tab:claims}: claim, evidence, causal or not, scale, and the claims we do \emph{not} make. \\
\S\ref{app:guide} Main-Paper Figure and Table Guide & Panel and column definitions for Figs.~\ref{fig:robust}--\ref{fig:conseqfull} and Tabs.~\ref{tab:ladder}a, \ref{tab:conseq}a and \ref{tab:conseq}b. & Tab.~\ref{tab:guide}: what every panel and column means. \\
\rowcolor{tblalt}\S\ref{app:addfigs} Additional Figures & Companion figures to the main results. & Fig.~\ref{fig:impact}: where the decision forms. Fig.~\ref{fig:ladder}: the read-out end to end. Fig.~\ref{fig:conseqfull}: the ten-panel version of Fig.~\ref{fig:conseq}. \\
\S\ref{app:roster} Model Registry and Dataset Sheet & The 64 models and 14 datasets, with roles, sizes, and provenance. & Tab.~\ref{tab:roster}: all 64 models by role. Tab.~\ref{tab:instr}: instruments, controls, compute. Tab.~\ref{tab:datasets}: the 8 pairwise benchmarks. Tab.~\ref{tab:aux}: the 6 auxiliary sets. \\
\rowcolor{tblalt}\S\ref{app:recipe} Deployable Read-Out Recipe & The read-out written out as runnable steps, with no weight updates. & Alg.~1: cache the residual stream, order-average, fit the probe. \\
\S\ref{app:eval} Extended Evaluator Results & Reading against retraining, abstention, cross-judge transfer, and which judge to deploy. & Tab.~\ref{tab:evaluator}: read vs.\ train at matched labels. Tab.~\ref{tab:er1}: abstention against the verdict margin. Tab.~\ref{tab:deploy}: judge re-ranking. \\
\rowcolor{tblalt}\S\ref{app:cost} Cost Accounting & Labels, gradient steps, parameters updated, and storage for each arm. & Tab.~\ref{tab:cost}: cost of every arm at the same label budget. \\
\S\ref{app:mech} Extended Mechanism & Which behavioral failures the mechanism accounts for, and the same story in reward models. & Tab.~\ref{tab:pathologies}: pathologies and internal repairs. Tab.~\ref{tab:coherence}: every measured gain ordered by $\mathrm{cf}$. \\
\rowcolor{tblalt}\S\ref{app:conseq} Consequential Experiments & Full per-cell tables behind \S\ref{sec:consequential}. & Tab.~\ref{tab:er8}: probe- against verdict-labeled DPO. \\
\S\ref{app:related} Extended Related Work & Our precise relation to each closest prior work, one by one. & \na \\
\rowcolor{tblalt}\S\ref{app:repro} Reproducibility Protocol & Seeds, determinism settings, and the numerical guards. & \na \\
\S\ref{app:limits} Limitations and Data Statement & What we do not claim, and where every dataset comes from. & \na \\
\end{longtable}}

\subsection{Additional Definitions}
\label{app:defs}
Two read-out quantities used in \S\ref{sec:anatomy} are defined here rather than in the
body. \textbf{Dimensional collapse} compares the effective rank of the internal score
representation with that of the emitted output score; we measure effective rank by the
participation ratio, an entropy-based effective-dimension measure of the eigenvalue
spectrum, so that a representation whose variance is spread over many directions scores
higher than one concentrated on a few. A \textbf{concept direction} $\hat u$ is the unit
weight vector of a linear probe fit on $h_\ell$ to predict a nuisance or target attribute
(position, length, language, quality); it is the direction we add in steering and project
out in ablation.

\subsection{Claim--Evidence Matrix}
\label{app:claims}
Table~\ref{tab:claims} lists every load-bearing claim in the main paper, the evidence behind it,
whether that evidence is causal or correlational, its scale, and the claims we deliberately do
\emph{not} make. It is intended to let a reader check the mapping from claim to evidence directly.

{\small
\setlength{\tabcolsep}{3pt}\renewcommand{\arraystretch}{1.12}
\begin{longtable}{@{}p{0.285\columnwidth}p{0.255\columnwidth}cp{0.145\columnwidth}p{0.115\columnwidth}@{}}
\caption{Claim--evidence matrix for the main paper.}
\label{tab:claims}\\
\toprule
\rowcolor{tblhead}\textbf{Claim} & \textbf{Evidence} & \textbf{Causal?} & \textbf{Scale} & \textbf{Where} \\
\midrule
\endfirsthead
\toprule
\rowcolor{tblhead}\textbf{Claim} & \textbf{Evidence} & \textbf{Causal?} & \textbf{Scale} & \textbf{Where} \\
\midrule
\endhead
\bottomrule
\endfoot
Preference is decodable from frozen activations when the verdict is wrong & CV probe, $.456\!\to\!.758/.846$ & no (L1) & 50 judges, LLMBar-adv & \S\ref{sec:gap} \\
\rowcolor{tblalt}That signal is not only surface & 9-dim residualization, length balancing, shuffle control & no (L1) & 50 judges & \S\ref{sec:gap} \\
The gap generalizes & ladder, $8/8$ benchmarks, $89\%$ of cells & no & $400$ cells & \S\ref{sec:gap} \\
\rowcolor{tblalt}\dots\ and is not an artifact of treating judges as independent & family-clustered bootstrap; $7/8$ CIs exclude $0$; positive in $21/21$ families & no & $21$ families & \S\ref{sec:gap} \\
Intervention changes the verdict from mid-stack & residual patching, steering & \hw{yes (L2)} & $41$ judges & \S\ref{sec:anatomy} \\
\rowcolor{tblalt}\dots\ before it is lens-decodable & paired per-judge test, $34/39$, $p\!=\!2{\times}10^{-7}$ & \hw{yes (L2)} & $39$ judges & \S\ref{sec:anatomy} \\
Attribution is an incomplete causal proxy & DLA vs.\ causal knockout, $\rho\!=\!.31$; $82\%$ indirect & \hw{yes (L2)} & $41$ judges & \S\ref{sec:attrcause} \\
\rowcolor{tblalt}Surface bias is a localizable circuit & head ablation $33$--$54\%$ vs.\ $\approx\!3\%$ random; subspace removal; transfer & \hw{yes (L2)} & $41$ judges & \S\ref{sec:anatomy} \\
$\mathrm{cf}$ predicts the gain & $\rho\!=\!.90$, exact perm.\ $p\!=\!.005$; partial $\rho\!=\!.90$; LOO $[.86,.96]$; held-out seed $\rho\!=\!.95$; cell-level $\hat\beta\!=\!.53$ & no & $8$ benchmarks, $400$ cells, $2$ seeds & \S\ref{sec:apriori} \\
\rowcolor{tblalt}$\mathrm{cf}$ predicts \emph{absence} & clean benchmarks $\delta_{\text{o}}\!\approx\!0$; held-out rubric format $\rho\!=\!-.006$ & no & $2$ benchmarks, $94$ judge-runs & \S\ref{sec:rubricnull} \\
Read-out beats output-level fixes at matched labels & $24$-cell \textsc{Budget} grid & no & $6$ judges $\times$ $4$ benchmarks & \S\ref{sec:consequential} \\
\rowcolor{tblalt}Read-out beats one tested full-FT recipe & matched-label comparison, $590$ shared items & no & $1$ judge (24B) & \S\ref{sec:consequential} \\
Probe labels train a better DPO policy & $5$ cells, gold-test accuracy & no & $5$ cells, $2$ policy sizes & \S\ref{sec:consequential} \\
\midrule
\rowcolor{hihot}\multicolumn{5}{@{}l}{\emph{Claims we do \textbf{not} make}} \\
\rowcolor{hihot}The judge \emph{used} the probe's direction & \na & \hh{L3, not shown} & \na & \S\ref{sec:setup} \\
\rowcolor{hihot}Probing beats fine-tuning in general & one untuned recipe & \hh{not shown} & \na & \S\ref{sec:consequential} \\
\rowcolor{hihot}The read-out transfers across benchmarks without labels & naive transfer fails ($-0.29$) & \hh{refuted} & \na & \S\ref{sec:apriori} \\
\rowcolor{hihot}$\mathrm{cf}$ predicts usefulness in every setting & one source of output-stage error & \hh{not shown} & \na & \S\ref{sec:rubricnull} \\
\end{longtable}}

\subsection{Guide to the Main-Paper Figures and Tables}
\label{app:guide}

Body captions are kept short. Table~\ref{tab:guide} gives the full reading of every panel
and column, with the statistics and controls behind each.

{\small
\setlength{\tabcolsep}{3pt}\renewcommand{\arraystretch}{1.14}
\begin{longtable}{@{}p{0.12\columnwidth}p{0.83\columnwidth}@{}}
\caption{Panel and column guide for the main-paper figures and tables.}
\label{tab:guide}\\
\toprule
\rowcolor{tblhead}\textbf{Panel} & \textbf{What it shows, and how it was computed} \\
\midrule
\endfirsthead
\toprule
\rowcolor{tblhead}\textbf{Panel} & \textbf{What it shows, and how it was computed} \\
\midrule
\endhead
\bottomrule
\endfoot
\multicolumn{2}{@{}l}{\emph{Fig.~\ref{fig:impact}: where the decision forms}} \\
\quad \na & A forward-pass residual patch places the causal decision at $0.58$ of depth, before the lens first decodes the verdict; per judge this ordering holds on $34/39$ judges (Fig.~\ref{fig:robust}d). On easy items the verdict is linearly readable without extra fitting; on adversarial items the untrained read-out is wrong at every depth, which is why the deployable reader is a \emph{trained} probe. \\
\midrule
\multicolumn{2}{@{}l}{\emph{Fig.~\ref{fig:ladder}: the read-out, end to end (50 judges $\times$ 8 benchmarks)}} \\
\quad (a) & A small-label lens probe (blue) and a residual-stream probe (green) exceed the stated verdict (rose) on all 8 benchmarks, mean $+0.12$, every one significant under FDR-BH; the margin is largest where the verdict collapses to chance. Training-free order-averaging (sand) beats raw on $7/8$. \\
\rowcolor{tblalt}\quad (b) & Selective prediction: abstaining on the lowest-confidence items raises accuracy monotonically as coverage falls. At $25\%$ coverage accuracy reaches $0.91$--$0.98$. \\
\quad (c) & Over all $400$ judge$\times$benchmark cells, absolute probe accuracy \emph{rises} with judge quality (slope $0.56$) and $89\%$ of cells sit above $y\!=\!x$, so the weak-judge gain is range compression rather than headroom. \\
\rowcolor{tblalt}\quad (d) & The surface-confound diagnostic \textbf{cf} orders the eight benchmarks by probe$-$verdict gain, $\rho\!=\!0.90$ (exact permutation $p\!=\!0.005$), leave-one-benchmark-out $\rho\!\in\![0.86,0.96]$. \\
\quad (e) & Given the same $256$ human labels the frozen probe recovers about half of the mean LoRA fine-tuning gain with zero gradient steps, and most of it on confounded llmbar (Table~\ref{tab:evaluator}). \\
\midrule
\multicolumn{2}{@{}l}{\emph{Fig.~\ref{fig:robust}: the gap and its statistics (main body)}} \\
\quad (a) & Reader hierarchy on LLMBar-adversarial over $50$ judges with $95\%$ bootstrap CIs, ordered stated verdict $\to$ surface-only $\to$ internal $\to$ surface-residualized internal $\to$ shuffled-label control. The hatched bars are the surface-controlled preference signal: the read-out after a 9-dimensional surface vector is regressed out. \\
\rowcolor{tblalt}\quad (b) & Per-benchmark $\Delta\!=\!$ probe $-$ raw, bootstrapped twice: over the $50$ judges (sand) and over the $21$ model families (blue), $20{,}000$ resamples each. Family clustering is the conservative unit of independence. \\
\quad (c) & cf vs.\ mean gain over the eight benchmarks, seed $0$ (circles) and the independent replication seed $1$ (triangles); the seed-$0$ diagnostic predicts seed-$1$ gains at $\rho\!=\!0.95$. \\
\rowcolor{tblalt}\quad (d) & Per-judge scatter of causal decision depth (residual patch) against lens decoding depth, $n\!=\!39$ judges with both measurements; points above the diagonal decode \emph{after} the intervention-sensitive onset. \\
\midrule
\multicolumn{2}{@{}l}{\emph{Fig.~\ref{fig:conseq}: four consequences (main body)}} \\
\quad (a) & The probe's paired advantage over each output-level fix: $+0.17$--$+0.21$ on confounded cells, $\approx\!0$ or negative on clean ones. \\
\rowcolor{tblalt}\quad (b) & Label-efficiency sweep on rb2: the downstream DPO gain grows as the labeled pool shrinks, peaking at $+0.029$ at $256$--$512$ labels. \\
\quad (c) & Ablating the position heads removes $44\%$ of the bias gap and the same circuit transfers to unseen data ($51\%$), while random-head ablation removes $\approx\!3\%$. \\
\rowcolor{tblalt}\quad (d) & Risk--coverage AUC of internal confidence vs.\ the verdict margin (Table~\ref{tab:er1}). \\
\midrule
\multicolumn{2}{@{}l}{\emph{Fig.~\ref{fig:conseqfull}: consequences, full 10 panels (24 cells, same labels)}} \\
\quad (a) & Mean accuracy of each reader over the 24 cells: probe $0.682$ vs.\ raw $0.568$, self-consistency $0.581$, calibrated prompt $0.617$. \\
\rowcolor{tblalt}\quad (b) & The probe's paired advantage over each output-level fix is $+0.17$--$+0.21$ on confounded cells and vanishes on clean ones. \\
\quad (c) & Probe-labeled preference data trains a better DPO policy on four of five cells; on LLMBar verdict-DPO falls below chance ($0.463$) while probe labels reach $0.787$. \\
\rowcolor{tblalt}\quad (d) & Label-efficiency sweep on rb2: the downstream gain grows as the labeled pool shrinks, peaking at $+0.029$ at $256$--$512$ labels. \\
\quad (e) & Against a full-parameter fine-tune of a 24B judge on the same labels the frozen probe wins overall ($0.753$ vs.\ $0.659$) and holds safety where full-FT regresses ($0.716\!\to\!0.569$ on pku-safe). \\
\rowcolor{tblalt}\quad (f) & Causal surface control: residualizing length and a 9-dimensional surface vector still leaves human-label-predictive signal well above the shuffled-label control. \\
\quad (g) & Ablating the position heads removes $44\%$ of the bias gap and the same circuit transfers to unseen data ($51\%$), while random-head ablation removes $\approx\!3\%$. \\
\rowcolor{tblalt}\quad (h) & Stacking the order-averaged final layer with mid-layer content beats either component alone (llmbar $0.738$ vs.\ $0.454$ / $0.582$). \\
\quad (i) & Risk--coverage AUC of internal confidence vs.\ the verdict margin: the learned internal reader gives a better abstention signal than the judge's own output margin (Table~\ref{tab:er1}). \\
\rowcolor{tblalt}\quad (j) & The measured gains from internals decline together with the same surface-confound diagnostic \textbf{cf} (Table~\ref{tab:coherence}). \\
\midrule
\multicolumn{2}{@{}l}{\emph{Table~\ref{tab:ladder}a: column definitions}} \\
\quad cf & Surface-confound diagnostic: cross-validated accuracy of a length-only probe on that benchmark, computed with no internal read-out and before any probe is fitted. \\
\rowcolor{tblalt}\quad raw & The judge's stated verdict. \\
\quad o-avg & Training-free order-average over both answer orders. \\
\rowcolor{tblalt}\quad probe & Small-label $L_2$-logistic probe on the logit-lens read-out. \\
\quad resid & The same probe on the residual stream (PCA-48). \\
\rowcolor{tblalt}\quad $\Delta_{\text{r}}$ & probe $-$ raw; all 8 significant under FDR-BH, both with the judge and with the model family as the resampling unit. \\
\quad $\delta_{\text{o}}$ & probe $-$ order-average: the probe's non-redundant value. \\
\rowcolor{tblalt}\quad mean & Eight-benchmark mean of each column on seed $0$ and on the independent replication seed $1$; the two seeds agree at cell level with $r\!\geq\!0.96$ for every reader. \\
\quad fam.\ CI$>$0 & Number of benchmarks whose family-clustered $95\%$ bootstrap CI for the gain over raw excludes zero, with the $21$ model families as the resampling unit. \\
\midrule
\multicolumn{2}{@{}l}{\emph{Table~\ref{tab:conseq}a: output-level fixes on the same labels}} \\
\rowcolor{tblalt}\quad cells & 24 judge$\times$benchmark cells (6 judges $\times$ 4 benchmarks), every arm scored on identical items with the same $256$ human labels. The read-out is the $41$-parameter logit-lens probe, as everywhere in \S\ref{sec:consequential}. \\
\quad self-cons. & $k\!=\!5$ samples at $T\!=\!0.7$, majority-voted. \\
\rowcolor{tblalt}\quad calibrated & A prompt instructing the judge to ignore length and position. \\
\quad mean & Accuracy averaged over all 24 cells. \\
\rowcolor{tblalt}\quad confounded & The probe's \emph{paired} advantage over that reader on the 12 confounded cells (llmbar, rb2). On the 12 clean cells (judgebench, arena) each of these advantages is $\approx0$ ($-0.05$ to $-0.01$), which is why the edge is a confounded-regime phenomenon. \\
\midrule
\multicolumn{2}{@{}l}{\emph{Table~\ref{tab:conseq}b: full fine-tuning on the same labels}} \\
\rowcolor{tblalt}\quad setup & A 24B judge (Mistral-Small-24B) is full-parameter fine-tuned on the verdict-format objective over $2\times8$ B200s with FSDP2; the frozen-judge probe is fit on the \emph{same} $3{,}349$ human labels. Both arms are scored on the same untouched 590-item test set, order-averaged. \\
\quad base & The judge before any fine-tuning. \\
\rowcolor{tblalt}\quad full-FT & After full-parameter fine-tuning. \\
\quad probe & The frozen judge read by the $41$-parameter logit-lens probe, with no weight updates. \\
\rowcolor{tblalt}\quad $n$ per row & Held-out items behind each row: rb2 $225$, llmbar $47$, judgebench $93$, pku-safe $225$; \emph{overall} $590$ and \emph{excl.\ pku-safe} $365$. \\
\quad pku-safe & The safety third of the pool (amber), where full-FT regresses $0.716\!\to\!0.569$ while the probe holds $0.707$. \\
\rowcolor{tblalt}\quad llmbar & The two arms tie at $0.660$; $n\!=\!47$ admits $\pm0.14$, so this cell is not separating. \\
\quad excl.\ pku & Overall accuracy with the safety split removed, isolating the non-safety comparison. \\
\end{longtable}}

\subsection{Additional Figures}
\label{app:addfigs}
\begin{figure}[h]
\centering
\includegraphics[width=0.60\textwidth]{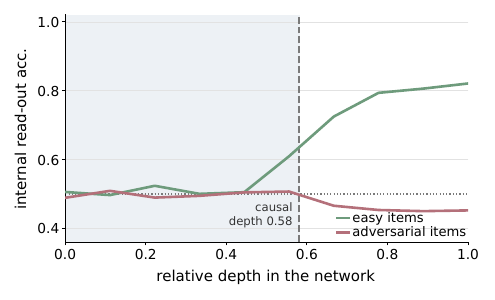}
\caption{Where the decision forms. A forward-pass residual patch
places the causal decision at $0.58$ of depth, before the lens decodes the verdict. On easy
items the verdict is linearly readable without extra fitting; on adversarial items the untrained
read-out is wrong at every depth, which is why the deployable reader is a \emph{trained}
probe. Per-judge pairing of the two depths: Fig.~\ref{fig:robust}d.}
\label{fig:impact}
\end{figure}

\begin{figure}[h]
\centering
\includegraphics[width=\textwidth]{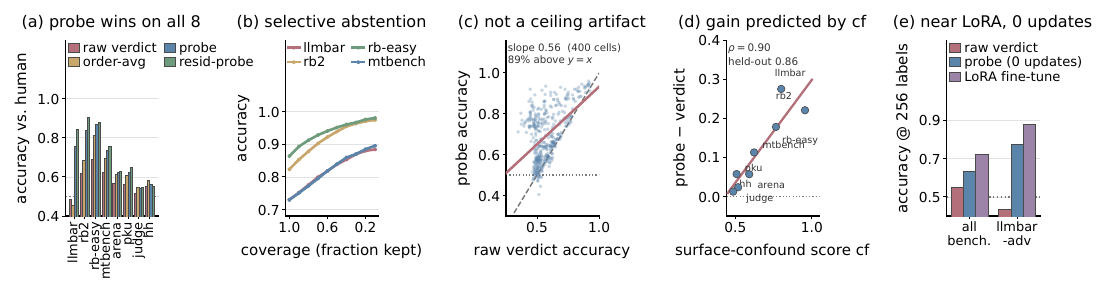}
\caption{The read-out end to end, over 50 judges and 8 benchmarks.
\textbf{(a)}~Both probes beat the stated verdict everywhere. \textbf{(b)}~Abstaining raises
accuracy monotonically. \textbf{(c)}~Not a ceiling artifact. \textbf{(d)}~The gain is
predicted by \textbf{cf} ($\rho\!=\!0.90$). \textbf{(e)}~At $256$ labels the frozen probe
recovers about half of the mean LoRA gain with zero gradient steps. Panel guide: Table~\ref{tab:guide}.}
\label{fig:ladder}
\end{figure}

This section collects two companion figures to the main results. Figure~\ref{fig:impact} shows
the depth evidence summarized per judge in Table~\ref{tab:ladder}b and Fig.~\ref{fig:robust}d;
Figure~\ref{fig:ladder} is the end-to-end evaluator panel whose headline rows appear in
Table~\ref{tab:ladder}.

\begin{figure}[h]
\centering
\includegraphics[width=\textwidth]{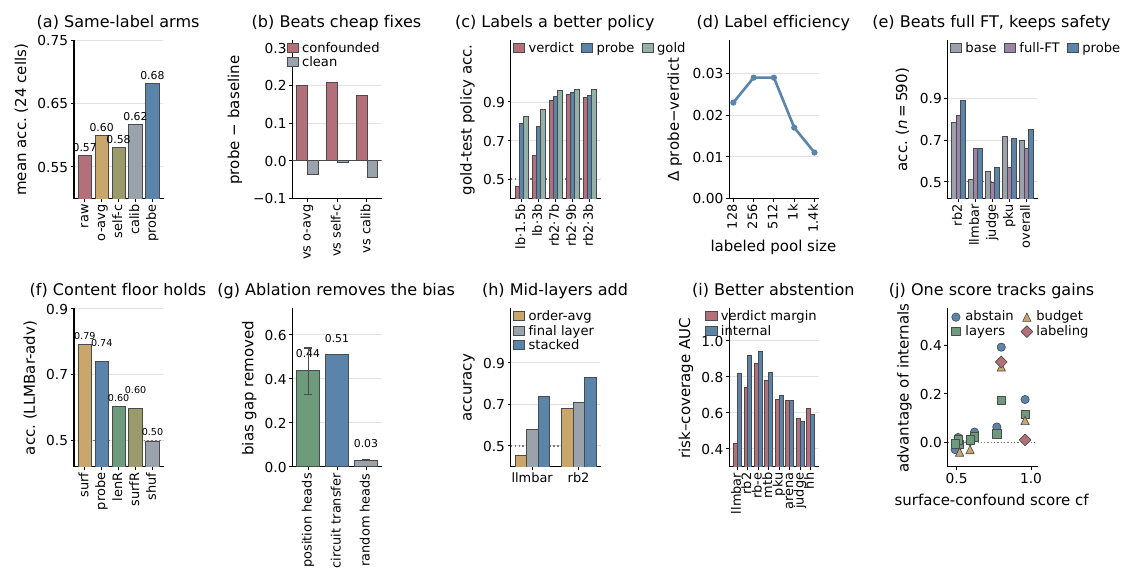}
\caption{Downstream consequences of the internal read-out, the full ten-panel version of Fig.~\ref{fig:conseq}. \textbf{(a)}~All arms given the same number of labels; probe best.
\textbf{(b)}~Its edge over output-level fixes is confined to confounded cells.
\textbf{(c)}~Probe labels train a better DPO policy on four of five cells. \textbf{(d)}~That
gain grows as labels get scarce. \textbf{(e)}~Against a \emph{full}-parameter fine-tune of a
24B judge, holding safety where that fine-tune regresses. \textbf{(f)}~Human-label-predictive signal
survives surface residualization (the surface-only bar is high by construction on this
adversarial set). \textbf{(g)}~Ablating the bias heads removes the bias, also on unseen data. \textbf{(h)}~Stacked
mid-layers beat either component. \textbf{(i)}~Internal confidence supports better abstention
than the judge's own verdict margin. \textbf{(j)}~The measured gains decline with $\mathrm{cf}$. Panel guide:
Table~\ref{tab:guide}.}
\label{fig:conseqfull}
\end{figure}

\subsection{Model Registry and Dataset Sheet}
\label{app:roster}

Table~\ref{tab:roster} lists all \textbf{64} registry models grouped by the role each plays.
Table~\ref{tab:datasets} is the dataset sheet: what each set contains, what the label is,
which claim it supports, and how it may be obtained. Table~\ref{tab:instr} inventories the
instruments and controls summarized in \S\ref{sec:setup}.

\begin{table}[h]
\centering\small
\setlength{\tabcolsep}{3pt}\renewcommand{\arraystretch}{1.12}
\begin{tabular}{@{}p{0.235\columnwidth}p{0.70\columnwidth}@{}}
\toprule
\rowcolor{tblhead}\textbf{Family} & \textbf{Models (sizes as loaded)} \\
\midrule
\rowcolor{hicold}\multicolumn{2}{@{}l}{\emph{Role A, the flagship adversarial panel: \textbf{50} generative judges.}} \\
\rowcolor{hicold}\multicolumn{2}{@{}l}{\emph{\hspace{1em}The forward-pass causal suite intervenes on the \textbf{41} of these that expose}} \\
\rowcolor{hicold}\multicolumn{2}{@{}l}{\emph{\hspace{1em}hookable decoder blocks within our memory budget; all 50 are read out.}} \\
Qwen2.5 & \texttt{0.5b, 1.5b, 3b, 7b, 14b, 32b} \\
\rowcolor{tblalt}Qwen3 & \texttt{8b, 14b, 30b-a3b, 32b} \\
Llama-3 & \texttt{3.2-1b, 3.1-8b, tulu-3-8b} \\
\rowcolor{tblalt}Gemma-2/3 & \texttt{2-2b, 2-9b, 2-27b, 3-12b, 3-27b} \\
Mistral & \texttt{7b-v0.3, ministral-8b, small-24b, mixtral-8x7b} \\
\rowcolor{tblalt}Yi / GLM / InternLM & \texttt{yi-1.5-9b, yi-1.5-34b, glm-4-9b, internlm2.5-7b, internlm2.5-20b} \\
OLMo-2 / SmolLM2 & \texttt{olmo2-7b, olmo2-13b, smollm2-360m, smollm2-1.7b} \\
\rowcolor{tblalt}Phi / Falcon / gpt-oss & \texttt{phi-3.5-mini, phi-4, falcon3-7b, gpt-oss-20b} \\
Reasoning & \texttt{deepseek-r1-\{llama-8b, qwen-7b, qwen-32b\}, qwq-32b} \\
\rowcolor{tblalt}Multilingual / Indic & \texttt{aya-23-8b, aya-expanse-\{8b,32b\}, command-r7b, sarvam-m} \\
Purpose-built judges & \texttt{prometheus-7b, prometheus-8x7b, judgelm-7b, offsetbias-8b, skywork-critic-8b, selene-mini-8b} \\
\midrule
\rowcolor{hicold}\multicolumn{2}{@{}l}{\emph{Role B, read-out-only large judges: \textbf{5}. Too large to intervene on}} \\
\rowcolor{hicold}\multicolumn{2}{@{}l}{\emph{\hspace{1em}under our memory budget; used for scale checks only.}} \\
70B and above & \texttt{llama-3.3-70b, qwen2.5-72b, command-r-plus, llama-3.1-405b, qwen3-235b} \\
\midrule
\rowcolor{hicold}\multicolumn{2}{@{}l}{\emph{Role C, scalar reward models: \textbf{9}. Verify the picture transfers from}} \\
\rowcolor{hicold}\multicolumn{2}{@{}l}{\emph{\hspace{1em}verdict-emitting judges to scalar heads; \textbf{5} are causally intervened.}} \\
Scalar RMs & \texttt{armorm-8b, skywork-rm-llama-8b, skywork-rm-gemma-27b, qrm-llama-8b, urm-llama-8b, internlm2-7b-reward, nemotron-70b-reward, grm-llama-3b, grm-gemma2-2b-rm} \\
\bottomrule
\end{tabular}
\caption{The full model registry, all 64 models by role. $\textbf{50}$ panel judges
$+$ $\textbf{5}$ read-out-only large judges $+$ $\textbf{9}$ scalar reward models $=$
$\textbf{64}$, spanning $0.36$B--$405$B and $20{+}$ families. The forward-pass causal suite
intervenes on \textbf{41} panel judges and \textbf{5} reward models; every other registry
entry is read out but not intervened on. $63$ of the $64$ are loadable locally.}
\label{tab:roster}
\end{table}

\begin{table}[h]
\centering\small
\setlength{\tabcolsep}{3pt}\renewcommand{\arraystretch}{1.12}
\begin{tabular}{@{}p{0.30\columnwidth}p{0.62\columnwidth}@{}}
\toprule
\rowcolor{tblhead}\textbf{Component} & \textbf{What we use} \\
\midrule
\rowcolor{hicold}\multicolumn{2}{@{}l}{\emph{Cached read-out surrogates}} \\
Lenses & logit lens; tuned lens (learned affine per layer) \\
\rowcolor{tblalt}Attribution & direct logit attribution over heads and MLPs \\
Patching & activation patching, path patching, causal scrubbing \\
\rowcolor{tblalt}Features & Gemma-Scope sparse autoencoders \\
Probes & $L_2$-logistic, 5-fold CV, shuffled-label control \\
\rowcolor{tblalt}Hooks & residual / attn-out / mlp-out at the decision position, \texttt{float32} \\
\midrule
\rowcolor{hicold}\multicolumn{2}{@{}l}{\emph{Forward-pass causal suite (the gold standard)}} \\
Steering & dose-response over $\alpha$, layer sweeps \\
\rowcolor{tblalt}Ablation & per-head knockout, subspace projection, random-head control \\
Decomposition & path and mediation decomposition (direct vs.\ indirect) \\
\rowcolor{tblalt}Transfer & debias-circuit transfer to unseen data \\
\midrule
\rowcolor{hicold}\multicolumn{2}{@{}l}{\emph{Statistical controls}} \\
Intervals & bootstrap $95\%$ CIs, 2000 iterations \\
\rowcolor{tblalt}Tests & paired sign / Wilcoxon, FDR-BH over sweeps and the ladder \\
Nulls & shuffled labels; random-component null check \\
\rowcolor{tblalt}Confound controls & length and position residualization; 9-dim surface vector \\
Seeds & causal suite $0/1/2$; flagship and ladder replicated at seed $1$ \\
\midrule
\rowcolor{hicold}\multicolumn{2}{@{}l}{\emph{Compute}} \\
Read-out & cached-activation backend, eight models at a time per node \\
\rowcolor{tblalt}Fine-tuning & $2\times8$ B200, FSDP2 (full-FT); single-GPU LoRA \\
\bottomrule
\end{tabular}
\caption{Instrument, control, and compute inventory for \S\ref{sec:setup}. Cached
surrogates are validated against the forward-pass suite, which changes the actual generation.}
\label{tab:instr}
\end{table}

\begin{table}[h]
\centering\small
\setlength{\tabcolsep}{3pt}\renewcommand{\arraystretch}{1.12}
\begin{tabular}{@{}p{0.245\columnwidth}p{0.30\columnwidth}p{0.35\columnwidth}@{}}
\toprule
\rowcolor{tblhead}\textbf{Dataset} & \textbf{Label / form} & \textbf{Role in this paper} \\
\midrule
\rowcolor{hihot}LLMBar (natural, \textbf{adversarial}) & human pairwise, surface decoupled from quality & flagship confounded regime; the verdict falls below chance here \\
\rowcolor{hihot}RewardBench\,2 & human pairwise & highest confound score ($\text{cf}\!=\!.96$); largest probe gain \\
RewardBench (easy) & human pairwise & mid-confound control \\
\rowcolor{tblalt}JudgeBench & human pairwise, length-neutral & the \emph{null}: no surface cue, no probe gain \\
MT-Bench (human) & human pairwise & human--human agreement $0.646$ (inter-annotator, not a label ceiling) \\
\rowcolor{tblalt}Chatbot Arena & $55$k human preferences & scale / in-the-wild preference \\
HH-RLHF & human helpful/harmless pairs & low-confound control \\
\rowcolor{tblalt}PKU-SafeRLHF & safety preference pairs & safety regression test for fine-tuning \\
HelpSteer2 & multi-attribute ratings & rubric / attribute decomposition \\
\rowcolor{tblalt}UltraFeedback & preference pairs & DPO training pool (\S\ref{sec:consequential}) \\
BiGGen-Bench & rubric scoring & rubric-format control for the collapse analyses \\
\rowcolor{tblalt}m-RewardBench & multilingual pairwise (English, Hindi, Bengali, Arabic, Spanish, Chinese) & cross-lingual geometry and transfer \\
\rowcolor{hiwin}Samiksha \citep{hamna2026samiksha} & \textbf{human} rubric scores, Hindi/Kannada/Malayalam & existing Indic rubric benchmark; supplies the out-of-domain test of \S\ref{sec:rubricnull} \\
\rowcolor{hiwin}Pariksha \citep{watts2024pariksha} & \textbf{human} rubric scores, 8 Indian languages (Bengali, Hindi, Kannada, Malayalam, Marathi, Odia, Tamil, Telugu) & existing Indic rubric benchmark; supplies the out-of-domain test of \S\ref{sec:rubricnull} \\
\bottomrule
\end{tabular}
\caption{Dataset sheet for all $14$ sets. The first eight are the pairwise-preference
benchmarks of the evaluator ladder; the last six are the auxiliary sets
(Table~\ref{tab:aux}). Amber marks the confounded regime where the internal read-out wins
most; green marks the two Indic rubric benchmarks. \textbf{All fourteen datasets are
existing, previously published resources; none is introduced by this work.} Twelve are public, as is Pariksha
\citep{watts2024pariksha}. Samiksha \citep{hamna2026samiksha} is not openly redistributable;
researchers who need it should contact its authors, who can share it under the terms it was
collected under. We release the schema, rubric definitions, language coverage, evaluation
prompts, and every result computed from it.}
\label{tab:datasets}
\end{table}

\paragraph{The 21 model families used for clustering.}
The family-clustered statistics of \S\ref{sec:gap} treat each base-model family as one
resampling unit. The $50$ panel judges fall into $\mathbf{21}$ such families:
Qwen2.5~($6$), Qwen3~($5$, incl.\ QwQ), Gemma~($5$), Mistral~($4$, incl.\ Ministral and
Mixtral), Aya/Cohere~($4$), DeepSeek-R1~($3$), Llama-3~($3$, incl.\ Tulu-3), Yi~($2$),
InternLM~($2$), OLMo-2~($2$), SmolLM2~($2$), Phi~($2$), Prometheus~($2$), and one each of
GLM, Falcon, gpt-oss, Sarvam, JudgeLM, OffsetBias, Skywork-Critic and Selene. Purpose-built
judges are counted by their \emph{own} identity rather than their base model, which is the
conservative choice here: merging them into their bases would reduce the number of clusters
and widen the family-clustered intervals only slightly, since each contributes a single judge.
Table~\ref{tab:roster} groups some of these families into shared rows for compactness; the
clustering uses the $21$ families listed above.

\paragraph{The six auxiliary sets.}
The evaluator ladder of \S\ref{sec:gap} requires pairwise-preference labels and a
length-only confound score, which only the eight ladder benchmarks supply. The remaining
six sets carry the mechanism, rubric, and multilingual analyses, and each appears in
$20$--$31$ separate experiments; Table~\ref{tab:aux} states what each contributes and the
headline number where one exists. The two Indic rubric benchmarks give the most direct
human-grounded check in the paper and the out-of-domain confirmation of
\S\ref{sec:rubricnull}.

\begin{table}[h]
\centering\small
\setlength{\tabcolsep}{3pt}\renewcommand{\arraystretch}{1.12}
\begin{tabular}{@{}p{0.20\columnwidth}cp{0.60\columnwidth}@{}}
\toprule
\rowcolor{tblhead}\textbf{Auxiliary set} & \textbf{exps.} & \textbf{Result reported in this paper} \\
\midrule
\rowcolor{hiwin}Samiksha & $31$ & Out-of-domain confirmation over $50$ judges ($n\!=\!80$): emitted score $\rho\!=\!-0.006$, held-out best layer $-0.005$, probe $+0.017$; best layer beats final on $52\%$ of judges, matching the $50\%$ the confound account predicts \\
\rowcolor{hiwin}Pariksha & $31$ & Out-of-domain confirmation over $44$ judges ($n\!=\!80$): $+0.013$ / $+0.003$ / $-0.053$; best layer beats final on $50\%$ of judges \\
BiGGen-Bench & $27$ & Rubric-format control: standalone rubric scoring commits at $0.9$--$1.0$ of depth, latest of any format \\
\rowcolor{tblalt}HelpSteer2 & $27$ & Multi-attribute rank-collapse measurements (\S\ref{sec:anatomy}) \\
m-RewardBench & $23$ & Cross-lingual direction geometry and transfer \\
\rowcolor{tblalt}UltraFeedback & $20$ & Preference pool for the DPO labeling experiment (\S\ref{sec:consequential}) \\
\bottomrule
\end{tabular}
\caption{The six auxiliary datasets. These do not enter the evaluator ladder,
which needs pairwise labels and a length-only confound score, but each is used in $20$--$31$
mechanism experiments. Green marks the two Indic rubric benchmarks, which supply the
out-of-domain confirmation of \S\ref{sec:rubricnull}.}
\label{tab:aux}
\end{table}

\subsection{The Deployable Read-Out Recipe}
\label{app:recipe}
The recipe below is the frozen-backbone reader referenced in \S\ref{sec:gap}: it first decides
\emph{whether} to read internals at all, then caches the residual stream over both answer orders,
order-averages to cancel position bias, featurizes with the logit lens (or a PCA of the raw
residual), and fits a small $L_2$-logistic probe. Only the probe is learned; no gradient step
touches the judge. Step~0 is the practical contribution of \S\ref{sec:apriori}: on a low-confound evaluation set the
free order-average already suffices and the remaining steps buy nothing. The threshold $\tau$ is a
heuristic read off the eight benchmarks of \S\ref{sec:apriori} rather than a calibrated decision
rule, and should be re-estimated on a new evaluation suite.

\begin{algobox}{Algorithm~1: Interpretability-native evaluator read-out}
\footnotesize
\setlength{\parindent}{0pt}
\textbf{\textsf{input}} frozen judge $J$; items $\{(q,a_A,a_B)\}$; labels $\mathcal{L}$; threshold $\tau$ \\
\textbf{\textsf{output}} preference $\hat{y}$ and confidence $c$ per item \\[2pt]
\textcolor{cslate!55}{\rule{\linewidth}{0.4pt}}\\[2pt]
\kw{0\; trigger}\ fit a surface-only probe on $\mathcal{L}$; $\mathrm{cf}\!\leftarrow$ its CV accuracy\\
\hspace*{3.4em}\kw{if}\ $\mathrm{cf}\!<\!\tau$\ \kw{then} return the order-averaged verdict and stop\\
\hspace*{3.4em}\textcolor{cgrey}{// $\tau\!=\!0.75$ is a rule of thumb from our eight benchmarks; recheck it on new settings}\\[1.5pt]
\kw{1\; read}\ \ \kw{for}\ $o\in\{AB,BA\}$: run $J$; cache residual $h_\ell$ at the\\
\hspace*{3.4em}decision position $\forall\ell$; \ $\hat{v}_\ell\!\leftarrow\!\mathrm{Unembed}(\mathrm{LN}_f(h_\ell))$\\[1.5pt]
\kw{2\; debias}\ order-average features over $o$ \ \textcolor{cgrey}{// cancels position bias}\\[1.5pt]
\kw{3\; featurize}\ $\phi\!\leftarrow\![\hat{v}_\ell]_\ell$ \emph{(lens)} \ \kw{or}\ $\phi\!\leftarrow\!\mathrm{PCA}_{48}(\Delta h_\ell)$ \emph{(resid.)}\\[1.5pt]
\kw{4\; fit}\ \ \ \ $g\!\leftarrow\!L_2$-logistic probe on $(\phi,y),\,\mathcal{L}$ \ \textcolor{cgrey}{// 5-fold CV + shuffle ctrl}\\[1.5pt]
\kw{5\; predict}\ $\hat{y}\!\leftarrow\!g(\phi)$; \ $c\!\leftarrow\!|g(\phi)-\tfrac12|$\\[1.5pt]
\kw{6\; audit}\ \ \kw{if}\ $c<\tau$\ \kw{then} abstain, route-to-human\\[2pt]
\textcolor{cslate!55}{\rule{\linewidth}{0.4pt}}
\end{algobox}

\subsection{Extended Evaluator Results}
\label{app:eval}

\paragraph{No weight updates, vs.\ training (main \S\ref{sec:consequential}).}
The main-body comparison gives every arm the \textbf{same 256 human labels}
(Table~\ref{tab:evaluator}); the per-benchmark results follow the confound
score: on confounded LLMBar the frozen probe and a LoRA/PEFT fine-tune both help
substantially (raw $0.437\to$ probe $0.774\to$ LoRA $0.881$), and on length-neutral JudgeBench
neither helps. On low-confound PKU-Safe the confound score predicts no internal advantage and
the probe is flat while LoRA helps, so the prediction holds in the negative direction too,
which is what makes cf predictive in both directions.

\begin{table}[h]
\centering\small
\setlength{\tabcolsep}{4pt}\renewcommand{\arraystretch}{1.1}
\begin{tabular}{@{}lcccc@{}}
\toprule
\rowcolor{tblhead}
\textbf{Reader (256-label)} & \textbf{Mean} & \textbf{llmbar} & \textbf{judge} & \textbf{pku} \\
\midrule
Raw verdict                 & $0.550$ & \hh{$0.437$} & $0.547$ & $0.667$ \\
\rowcolor{hiwin}Probe (0 updates) & $0.634$ & $0.774$ & $0.537$ & $0.592$ \\
LoRA / PEFT fine-tune       & $\mathbf{0.724}$ & $0.881$ & $0.537$ & $0.753$ \\
\bottomrule
\end{tabular}
\caption{Reading the \emph{same} frozen judge vs.\ training, given the same 256 human labels
(4 judges $\times$ 3 benchmarks). The no-update probe (green) recovers about half of the mean
LoRA/PEFT gain and most of it on confounded llmbar (amber); against a \emph{full}-parameter
fine-tune the probe wins outright (main Table~\ref{tab:conseq}b).}
\label{tab:evaluator}
\end{table}

\paragraph{Auditor and selective prediction.}
Two cross-model auditors must be distinguished, because they support different claims. The
\textbf{internals-only} auditor, a probe fitted on the other $49$ judges' lens features and
applied to a held-out judge with \emph{none} of that judge's labels, raises human agreement over
the held-out judge's own verdict by $+0.077$ on LLMBar (beating it for $78\%$ of judges) and is
flat to slightly negative on JudgeBench ($-0.007$) and RB2 ($-0.031$). This is the number we
report in the body, and its sign pattern follows $\mathrm{cf}$ exactly as \S\ref{sec:apriori}
predicts. A \textbf{combined internal-plus-length} auditor reaches a much larger $+0.322$
(LLMBar) / $+0.273$ (RB2), but on these two sets the length feature is itself a within-benchmark
oracle ($\mathrm{cf}\!=\!0.80$ / $0.96$), so that figure mostly measures the surface confound
rather than transferred internal structure; we therefore do not use it to support any claim about
internals. Integrating the
internal-confidence risk--coverage curve beats a no-selection baseline on all $8$
benchmarks (AUC lift $+0.02$ to $+0.09$; Fig.~\ref{fig:conseq}d), and at $25\%$ coverage
accuracy reaches $0.91$--$0.98$; a route-to-human policy auto-decides $100\%$ (RB2) / $88\%$
(rb-easy) at a $\geq90\%$ precision point, and correctly declines to auto-decide any of the
hard subjective sets, so it abstains on the sets where automation is least reliable.

\paragraph{What transfers, and what needs labels.}
\emph{Across judges}, the internals-only zero-label auditor transfers to an unseen judge
($+0.08$ LLMBar, beating its verdict for $78\%$ of judges) and not to the clean benchmarks. \emph{Across benchmarks}, a probe applied
naively does not transfer (llmbar$\to$rb2 $-0.29$; leave-one-benchmark-out beats the verdict
on $1/8$); only a probe \emph{pooled} over datasets transfers leave-one-dataset-out ($+0.03$,
CI $[+0.016,+0.076]$). The reader is thus judge-portable and distribution-specific:
deployment needs a small label set matched to the evaluation distribution, which is the
usual case, since you know your evaluation set in advance.

\paragraph{Abstain: verdict-margin selective baseline (full).}
Risk--coverage AUC of internal-probe confidence vs.\ the verdict margin over 50 judges/bench
(Table~\ref{tab:er1}): internals beat the trivial verdict-confidence baseline by $+0.39$
(LLMBar) / $+0.18$ (RB2) and converge on length-neutral benches.

\begin{table}[h]
\centering\small\setlength{\tabcolsep}{4pt}\renewcommand{\arraystretch}{1.1}
\begin{tabular}{@{}lccc@{}}
\toprule
\rowcolor{tblhead}\textbf{Bench} & \textbf{AUC int.} & \textbf{AUC verd.} & \textbf{$\Delta$ / \% win} \\
\midrule
\rowcolor{hihot}llmbar & 0.822 & 0.431 & $+.391$ / 100\% \\
\rowcolor{hihot}rb2 & 0.921 & 0.744 & $+.176$ / 98\% \\
rb-easy & 0.939 & 0.876 & $+.063$ / 74\% \\
mtbench & 0.823 & 0.780 & $+.042$ / 48\% \\
pku-safe & 0.697 & 0.677 & $+.020$ / 56\% \\
arena & 0.671 & 0.667 & $+.004$ / 32\% \\
judgebench & 0.555 & 0.567 & $-.013$ / 30\% \\
hh-rlhf & 0.592 & 0.622 & $-.030$ / 22\% \\
\bottomrule
\end{tabular}
\caption{Risk--coverage AUC: internal confidence vs.\ verdict margin. On llmbar the
verdict-margin AUC is $0.431$ (below chance): the judge's own confidence is
\emph{anti}-correlated with correctness.}
\label{tab:er1}
\end{table}

\paragraph{Deployment re-ranking (main \S\ref{sec:apriori}).}
Table~\ref{tab:deploy} gives the per-benchmark comparison behind the claim that reading
internals changes which judge you would deploy.

\begin{table}[h]
\centering\small
\setlength{\tabcolsep}{2.5pt}\renewcommand{\arraystretch}{1.06}
\begin{tabular}{@{}lcll@{}}
\toprule
\rowcolor{tblhead}\textbf{Bench.} & \textbf{$\rho$} & \textbf{by verdict} & \textbf{by internals} \\
\midrule
\rowcolor{hihot}rewardbench2 & .58 & skywork-crit-8b & \emph{unchanged} \\
\rowcolor{hihot}llmbar-adv & \hh{.20} & skywork-crit-8b & mistral-sm-24b \\
rb-easy & .71 & qwen2.5-14b & qwen2.5-32b \\
\rowcolor{tblalt}mtbench & .54 & phi-4 & ds-r1-qwen-32b \\
arena & .71 & mistral-sm-24b & gemma-3-27b \\
\rowcolor{tblalt}judgebench & \hh{.32} & qwen2.5-32b & ds-r1-qwen-32b \\
pku-safe & .74 & gemma-2-27b & qwen2.5-32b \\
\rowcolor{tblalt}hh-rlhf & .59 & qwen2.5-32b & offsetbias-8b \\
\midrule
\rowcolor{hiwin}\textbf{mean} & \textbf{.55} & \multicolumn{2}{r}{\textbf{changed on 7 of 8}} \\
\bottomrule
\end{tabular}
\caption{Deployment re-ranking: Spearman $\rho$ between ranking the 50 judges by
emitted verdict and by internal read-out, with the top-ranked judge under each. Rows follow
the confound ordering of Table~\ref{tab:ladder}. The best judge to deploy changes on 7 of 8
benchmarks and the two rankings agree only moderately (mean $\rho\!=\!0.55$). Agreement is
lowest on LLMBar-adversarial ($0.20$), where the emitted verdict is least trustworthy, and
also low on JudgeBench ($0.32$), where near-ties among judges make the ranking unstable
under either view (amber). RewardBench-2 is the one benchmark whose top judge is stable.
Names abbreviated (ds-r1 $=$ deepseek-r1, sm $=$ small, crit $=$ critic).}
\label{tab:deploy}
\end{table}

\subsection{Cost Accounting}
\label{app:cost}
The body claims the read-out is cheaper than fine-tuning. Table~\ref{tab:cost} makes that
precise for the comparison of \S\ref{sec:consequential}, in which every arm
sees the same $3{,}349$ labels and the same held-out $590$ items. We report an analytic
accounting (labels, gradient steps, parameters updated, stored artifact, forward passes per
item) rather than wall-clock times, which we did not instrument uniformly across arms. The
parameter and gradient-step counts are exact and hardware-independent, which is the
comparison that transfers across sites; wall-clock would be specific to our cluster.

The headline asymmetry is the parameter count. On Mistral-Small-24B ($40$ layers, model
width $5120$) the lens feature vector has one entry per layer, so the learned reader is $40$
weights and a bias: \textbf{$41$ parameters against $23.6$ billion}, and a $164$-byte
artifact against a $94$\,GB checkpoint. The residual-stream reader adds a $48$-component PCA
basis ($\approx\!1$\,MB) and $49$ probe parameters. Both readers require two forward passes
per item, one per answer order, which is the same cost as the training-free order-average
baseline and less than self-consistency at $k\!=\!5$.

\begin{table}[h]
\centering\small
\setlength{\tabcolsep}{3pt}\renewcommand{\arraystretch}{1.1}
\begin{tabular}{@{}lccrl@{}}
\toprule
\rowcolor{tblhead}\textbf{Arm} & \textbf{fwd/item} & \textbf{grad.\ steps} & \textbf{params upd.} & \textbf{stored} \\
\midrule
raw verdict & $1$ & $0$ & $0$ & \na \\
\rowcolor{tblalt}order-average & $2$ & $0$ & $0$ & \na \\
self-consistency ($k{=}5$) & $5$ & $0$ & $0$ & \na \\
\rowcolor{tblalt}calibrated prompt & $1$ & $0$ & $0$ & \na \\
\rowcolor{hiwin}lens probe & $2$ & $0$ & $41$ & $164$\,B \\
\rowcolor{hiwin}residual-stream probe & $2$ & $0$ & $49$ & $\approx\!1$\,MB \\
LoRA ($r{=}16$, $\alpha{=}32$) & $2$ & yes & adapter & adapter ckpt \\
\rowcolor{hihot}full-parameter FT (24B) & $2$ & 3 epochs & $23.6$\,B & $94$\,GB \\
\bottomrule
\end{tabular}
\caption{Cost of each arm when all arms are given the same $3{,}349$ human labels in
\S\ref{sec:consequential}. Parameter counts are for Mistral-Small-24B ($40$ layers, width
$5120$); the lens probe learns one weight per layer plus a bias. Green marks the arms that
touch no model weights; amber the arm they are compared against. This is an analytic
accounting, not a wall-clock benchmark.}
\label{tab:cost}
\end{table}

\subsection{Extended Mechanism}
\label{app:mech}

\paragraph{Behavioral pathologies the mechanism explains.}
Table~\ref{tab:pathologies} lists the observable failure modes the mechanism accounts for,
and two internal \emph{repairs} (rubric decompression; calibration).

\begin{table}[h]
\centering\small\setlength{\tabcolsep}{3.5pt}\renewcommand{\arraystretch}{1.1}
\begin{tabular}{@{}lcc@{}}
\toprule
\rowcolor{tblhead}\textbf{Property (behavioral probe)} & \textbf{Measure} & \textbf{Value} \\
\midrule
\rowcolor{hihot}\multicolumn{3}{@{}l}{\emph{Bias \& safety pathologies}} \\
\quad Position-bias flip rate & swap-flip frac. & $0.95\text{--}1.0$ \\
\quad Swap non-invariance & $1-$agreement & up to $1.0$ \\
\quad Eval-awareness shift & lens $\sigma$ & $2.0\text{--}5.6$ \\
\quad Self-preference (cross-gen) & win rate & $0.553$ \\
\quad\quad scaling with size & Spearman & $+0.63$ \\
\quad Contamination inflation & lens $\sigma$ & $+0.41$ \\
\quad CoT pre-commit & frac.\ before & $\approx\!0.80$ \\
\rowcolor{hicold}\multicolumn{3}{@{}l}{\emph{Reassuring separability}} \\
\quad Safety $\perp$ quality & $1-$overlap & \hw{$0.98$} \\
\quad Reward-hacking (spurious) & $R^2$ & $\approx\!-0.18$ \\
\rowcolor{hiwin}\multicolumn{3}{@{}l}{\emph{Interpretability-native repairs}} \\
\quad Rubric decompression & eff.\ rank & $2.95\!\to\!3.5$ \\
\quad Calibration (confidence) & Spearman & $-0.108\!\to\!+0.072$ \\
\quad Calibration error & ECE & $0.405\!\to\!0.262$ \\
\bottomrule
\end{tabular}
\caption{Behavioral pathologies the mechanism explains, with operational units. The last
block are internal repairs.}
\label{tab:pathologies}
\end{table}

\paragraph{Reward models and complementarity.}
Scalar RMs also form the reward mid-stack, where a mid layer reads it $+0.2$--$0.3$ better
than the scalar head (final $0.055\to$ best $0.355$; divergence $+0.33$). Per-judge signals
are complementary: a naive mean fails, but a sign-aligned ensemble beats the best single
judge ($0.642$ vs.\ $0.438$).

\begin{table}[h]
\centering\small
\setlength{\tabcolsep}{2.4pt}\renewcommand{\arraystretch}{1.08}
\begin{tabular}{@{}lcccccc@{}}
\toprule
\rowcolor{tblhead}\textbf{Bench.} & \textbf{cf} & \textsc{Measure} & \textsc{Abstain} & \textsc{Layers} & \textsc{Budget} & \textsc{Label} \\
\rowcolor{tblhead} & \emph{diagnostic} & \emph{acc.} & \emph{AUC} & \emph{content} & \emph{budget} & \emph{DPO} \\
\midrule
\rowcolor{hihot}rewardbench2 & .96 & \hw{$+.22$} & \hw{$+.18$} & \hw{$+.12$} & \hw{$+.09$} & $+.01$ \\
\rowcolor{hihot}llmbar-adv & .80 & \hw{$+.28$} & \hw{$+.39$} & \hw{$+.17$} & \hw{$+.31$} & \hw{$+.33$} \\
rb-easy & .77 & $+.18$ & $+.06$ & $+.04$ & \na & \na \\
\rowcolor{tblalt}mtbench & .62 & $+.11$ & $+.04$ & $+.02$ & \na & \na \\
arena & .59 & $+.06$ & $+.00$ & $+.01$ & $-.03$ & \na \\
\rowcolor{tblalt}judgebench & .52 & $+.02$ & $-.01$ & $-.01$ & $-.04$ & \na \\
pku-safe & .51 & $+.06$ & $+.02$ & $+.01$ & \na & \na \\
hh-rlhf & .49 & $+.01$ & $-.03$ & $-.01$ & \na & \na \\
\bottomrule
\end{tabular}
\caption{One score tracks the measured gains. The measurement gain
(\textsc{Measure}, probe$-$verdict accuracy), selective-prediction advantage
(\textsc{Abstain}, risk--coverage AUC $\Delta$), mid-layer content (\textsc{Layers}),
same-label advantage over cheap fixes (\textsc{Budget}), and downstream DPO gain
(\textsc{Label}) all track the surface-confound diagnostic \textbf{cf} and shrink to
$\approx0$ on length-neutral sets. Amber rows are the confounded regime where every column
peaks (green); the three full columns span all $8$ benchmarks, while
\textsc{Budget}/\textsc{Label} were run on 4- and 2-benchmark grids
($\na=$ not in that grid).}
\label{tab:coherence}
\end{table}

\paragraph{Cross-experiment coherence.}
Table~\ref{tab:coherence} ranks all eight benchmarks by the
confound score and shows the selective (\textsc{Abstain}), content (\textsc{Layers}), same-label (\textsc{Budget}), and
downstream (\textsc{Label}) advantages declining together; \textsc{Label}'s cells are reported per
bench$\cdot$judge in Table~\ref{tab:er8} below.

\subsection{Consequential Experiments: Full Tables}
\label{app:conseq}

\paragraph{Labeling downstream RLHF (all cells).}
Table~\ref{tab:er8} lists the five headline cells (gold-test accuracy) and the rb2$\cdot$3B
label-efficiency sweep. Probe labels win on four of the five cells; the gain scales with
confound and grows as labels get scarce. The rb2$\cdot$q7b$\cdot$3B cell is the single null:
the row reports the full-pool point of the label-efficiency sweep ($+0.011$), while a
separate run of the same configuration returns $0.000$, so we count that cell as level
rather than a win.

\begin{table}[h]
\centering\small\setlength{\tabcolsep}{3.5pt}\renewcommand{\arraystretch}{1.1}
\begin{tabular}{@{}llcccc@{}}
\toprule
\rowcolor{tblhead}\textbf{bench$\cdot$judge} & \textbf{pol.} & \textbf{verd.} & \textbf{probe} & \textbf{gold} & \textbf{$\Delta$} \\
\midrule
\rowcolor{hihot}llmbar$\cdot$q7b & 1.5b & 0.463 & \hw{0.787} & 0.825 & $+.325$ \\
\rowcolor{hihot}llmbar$\cdot$q7b & 3b & 0.625 & \hw{0.775} & 0.863 & $+.150$ \\
rb2$\cdot$q7b & 1.5b & 0.906 & \hw{0.931} & 0.960 & $+.026$ \\
rb2$\cdot$g9b & 1.5b & 0.940 & \hw{0.951} & 0.963 & $+.011$ \\
rb2$\cdot$q7b & 3b & 0.923 & \hw{0.934} & 0.963 & $+.011$ \\
\midrule
\multicolumn{6}{@{}l}{\emph{rb2$\cdot$3b label-efficiency (pool $\to$ $\Delta$ probe$-$verd.)}}\\
\quad 128 & & 0.880 & 0.903 & & $+.023$ \\
\quad 256 & & 0.900 & 0.929 & & \hw{$+.029$} \\
\quad 512 & & 0.923 & 0.951 & & \hw{$+.029$} \\
\quad 1024 & & 0.923 & 0.940 & & $+.017$ \\
\quad 1387 & & 0.923 & 0.934 & & $+.011$ \\
\bottomrule
\end{tabular}
\caption{Probe- vs.\ verdict-labeled DPO, gold-test accuracy. The gain is largest on
confounded llmbar (verdict-DPO below chance) and, on rb2, grows as the labeled pool shrinks.}
\label{tab:er8}
\end{table}

\paragraph{Layers stacked reader.}
The full-lens probe is the stack of the order-averaged verdict (final layer) $\oplus$ the
mid-layer content. It dominates both components on high-confound benches: llmbar $0.738$ vs.\
order-avg $0.454$ / final-only $0.582$ (100\% of judges); rb2 $0.828$ vs.\ $0.678$ / $0.707$
(96\%); neutral on clean benches.

\paragraph{Surface full 9-dim surface control.}
On the four aligned benches the preference signal survives residualizing the full 9-dim surface
vector; \emph{surface-resid} $\approx$ \emph{length-resid} on all four (within $0.005$), so
length is the dominant surface confound and the extra eight features (markdown, lists, code,
headers, bold, sentence length, digits) remove nothing more. On rb2 the surface vector alone
predicts the label at $0.980$ yet residual content still survives ($+0.086$ over shuffle).

\subsection{Extended Related Work}
\label{app:related}

Our work sits at the intersection of three threads; Table~\ref{tab:related} summarizes the positioning.

\paragraph{(i) Evaluation.}
LLM judges \citep{zheng2023judging,liu2023geval} and reward models
\citep{christiano2017deep,ouyang2022training} are audited \emph{behaviorally} through
benchmarks \citep{lambert2025rewardbench,tan2025judgebench,gureja2025mrewardbench} that
report agreement with human preference but say nothing about the computation producing it.
Known failure modes include position bias \citep{wang2024fair}, length/verbosity bias
\citep{dubois2024length,saito2023verbosity}, self-preference
\citep{panickssery2024self}, sycophancy \citep{sharma2024sycophancy}, and
adversarial-surface failure \citep{zeng2024llmbar}. Such confounds also distort measurement beyond
judging: in multilingual agent evaluation, trace length and chance agreement can each flip a
conclusion, and a single trace-extraction step, with the model unchanged, manufactured an apparent
failure \citep{mukherjee2026actions}. Remedies act at the \emph{output}:
preference optimization \citep{rafailov2023dpo}, multi-objective reward heads
\citep{wang2024armorm}, and cheap debiases such as swap-and-average or self-consistency
\citep{wang2023selfconsistency}. Reward hacking \citep{skalse2022defining} frames the risk of
optimizing such a proxy. None of these opens the evaluator.

\paragraph{(ii) Mechanistic interpretability.}
We use the logit \citep{nostalgebraist2020logitlens} and tuned \citep{belrose2023tunedlens}
lenses, direct logit attribution \citep{elhage2021mathematical}, activation/path patching and
causal tracing \citep{meng2022rome,wang2023ioi,goldowskydill2023localizing}, causal scrubbing
\citep{chan2022causalscrubbing} and mediation analysis \citep{vig2020causal}, sparse
autoencoders
\citep{bricken2023monosemanticity,huben2024sparse,templeton2024scaling,lieberum2024gemmascope},
and linear probing \citep{alain2017probing,belinkov2022probing} under superposition
\citep{elhage2022superposition}. We confirm at roster scale the caution that attribution need
not predict causal effect \citep{zhang2024patching}, and quantify \emph{why}: the effect is
mostly MLP-mediated, which a direct-path surrogate cannot see.

\paragraph{(iii) Reading internals for evaluation.}
Probes can detect latent knowledge that the output does not state
\citep{burns2023discovering}, and selective prediction supplies principled abstention
\citep{elyaniv2010foundations,guo2017calibration}. Four recent lines apply this to evaluators
directly, and we state our relation to each precisely.

\citet{maiya2025preference} train supervised and unsupervised linear probes on contrastive
prompt pairs and show they extract preferences more accurately than generation-based judgment
across four model families and six datasets, generalize under domain shift, and can match
fine-tuned evaluators at equal data. This establishes the descriptive claim (L1) that our
\S\ref{sec:gap} also observes, at larger roster scale and under a harder adversarial split.
\emph{We do not claim that observation as a contribution.} What we add is orthogonal: a causal
account of \emph{where} the evaluator's decision and biases are established
(\S\ref{sec:anatomy}) and a pre-read-out diagnostic for \emph{when} the read-out pays off
(\S\ref{sec:apriori}), including a case where it does not.

\citet{lai2025lager} (LAGER) improve point-wise judge--human correlation on Flask, HelpSteer,
and BIGGen by aggregating score-token logits across layers with the backbone frozen, reporting
that mid-to-upper layers align better with human scores than the final layer, independent
support for the mid-stack picture of \S\ref{sec:anatomy}. We differ in target and method: they
aggregate logits into a single point-wise score with no training, whereas we intervene causally
and work mainly in the pairwise setting where surface cues separate candidates.
\S\ref{sec:rubricnull} discusses the apparent sign disagreement between their point-wise gains
and our absolute-rubric null, and proposes the reconciliation that internal read-out helps in
proportion to the error the output stage introduces, of which surface confounding is one source
and score-token miscalibration another.

\citet{girrbach2025latent} derive scalar ratings from three latent signals (probability-weighted
scores over integer ratings, verifier-style ``yes'' probabilities, and probes on activations at
the rating position) and show they improve reference-free evaluation, Best-of-$N$ selection,
and routing, with multi-teacher distillation proposed as a further use. Their finding that probes recover signal specifically
when output logits are miscalibrated is the closest prior statement of a \emph{conditional}
account, and is consistent with ours; we make the condition measurable before the read-out is
fitted for pairwise preference and validate it on a held-out replication seed. \citet{li2026inspector} frame evaluation as
representation probing in small models (INSPECTOR), predicting aspect-level scores from
intermediate features; their aim is efficiency, replacing a large judge, rather than
diagnosing a large judge's failures.

Finally, concurrent with this work, \citet{feldhus2026judgecircuits} apply position-aware edge
attribution patching to judges in three model families and identify a sparse ``latent
evaluator'' sub-graph in mid-to-late MLPs that is shared across judgment tasks and feeds
fragile, format-specific output branches. That is the closest mechanistic result to ours and it
agrees with two of our findings on independent evidence: the evaluation computation is
established before the output stage, and the MLP pathway carries much of it. Their question is
why scores change \emph{across output formats}; ours is whether the human-labeled preference
survives the output stage at all, measured with forward-pass interventions in $41$ judges and
connected to evaluation and preference-learning outcomes.

Unlike prior interpretability work, which targets \emph{capabilities}, we target the
\emph{evaluator} itself, at roster scale, and convert the findings into a deployed reader whose
labels demonstrably improve downstream training.

\paragraph{Operationalizing causal decision depth.}
For each judge we sweep layers on a fixed grid of $10$ normalized depths. At each depth we patch
the residual stream at the decision position from a counterfactual run (the same item with the
two answers swapped) and re-run the forward pass to completion, recording whether the
\emph{emitted} verdict changes. A depth counts as flipping the verdict if it does so on a
majority of the item pairs for that judge ($>\!50\%$ of $8$--$12$ paired items, depending on the
judge's memory budget), and \textbf{causal decision depth} is the shallowest such depth on the
grid. We take the shallowest rather than the peak because the quantity of interest is the
\emph{onset} of causal sufficiency; because it is an onset on a coarse grid, it is stable to seed
(reported across seeds $0/1/2$) but should not be read to a precision finer than the grid
spacing. Judges for which no depth reaches the majority criterion are excluded from the depth
statistics, which is why the per-judge depth comparison of \S\ref{sec:anatomy} uses $39$ of the
$41$ causally intervened judges.

\subsection{Reproducibility Protocol}
\label{app:repro}
All runs are deterministic (\texttt{PYTHONHASHSEED}=0, offline caches). Model seeds are
$0/1/2$: the causal suite runs at all three and is seed-stable, the flagship adversarial
study and the evaluator ladder are reported at seed $0$ and replicate at seed $1$ within
$\pm0.035$. The fixed seed $42$ is used only for dataset splitting and stratified
subsampling, never for model execution. Numerical pipelines are guarded by unit checks
against a random-component null.

The full-parameter fine-tune of \S\ref{sec:consequential} uses Mistral-Small-24B-Instruct
trained on the verdict-format objective with AdamW at learning rate $6\!\times\!10^{-6}$,
cosine schedule, warmup ratio $0.03$, weight decay $0$, gradient clipping $1.0$, $3$ epochs,
micro-batch $2$ with gradient accumulation $2$ (effective batch $4$), sequence length
$2048$ with sample packing, bf16, gradient checkpointing and flash attention, sharded with
FSDP2 over $2\times8$ B200s. These are the framework defaults; we ran no hyperparameter
search, and the shipped configuration is released with the code. The LoRA baseline uses
$r\!=\!16$, $\alpha\!=\!32$, dropout $0.05$. Probes are $L_2$-regularized logistic
regressions under 5-fold cross-validation with shuffled-label controls.

We release the code, the model registry, per-experiment result artifacts, the evaluation
prompts, the 24-cell \textsc{Budget} grid, and the multi-node full-FT configuration.
Twelve of the fourteen
datasets are public, as is Pariksha \citep{watts2024pariksha}. Samiksha
\citep{hamna2026samiksha} is a restricted human-annotated corpus that we cannot
redistribute: its schema, rubric definitions, language coverage, evaluation prompts, and all
results computed from it are released, and the corpus itself is available from its
custodians under the data-use terms under which it was collected. Every model and dataset is
used under the license and terms of use set by its original creators; we redistribute no
model weights and no underlying corpus.

\subsection{Extended Limitations and Data Statement}
\label{app:limits}

\textbf{Decodability is not causal use.} Our probes establish that human-labeled preference is
recoverable from frozen activations (L1) and our interventions establish that specific
components causally change the emitted verdict (L2); neither establishes that the judge used
the probe's direction to form its own verdict (L3, \S\ref{sec:setup}). We report a result
consistent with that separation, since causal decision depth does not predict probe gain, and avoid
``the judge knows'' phrasings throughout. Relatedly, our evidence combines forward-pass
interventions, run on the registry subset exposing hookable decoder blocks within our compute
budget, with cached read-out surrogates run across the full registry and validated against
those interventions (median $\rho\!=\!0.78$, positive for $40/41$ judges); we report the
agreement between the two so readers can weigh each.

\textbf{Scope of the read-out and the diagnostic.} The read-out is fitted per evaluation
distribution and does not transfer naively across benchmarks; $\mathrm{cf}$ requires a small
labeled sample of that distribution, so it is pre-read-out, not label-free. It captures one
concrete source of output-stage error, namely surface confounding, and is not a general predictor of
every setting in which internals might help; \S\ref{sec:rubricnull} discusses one such setting
reported in prior work. Cost is parameter- and training-efficient rather than uniformly cheap:
both readers need two forward passes per item, and residual-stream features must be cached.

\textbf{Baselines and labels.} Fine-tuning baselines use standard published configurations rather
than a per-benchmark hyperparameter search, so they are representative reference points rather
than upper bounds, and we explicitly do not claim that probing outperforms a well-tuned
fine-tuning procedure in general. Human labels are noisy annotator-derived targets, not ground
truth; where we report
inter-annotator agreement it characterizes label consistency and does not bound the accuracy
attainable against one annotator-derived label set.

\textbf{Data statement.} All fourteen datasets are existing, previously published resources and
none is introduced by this work. Thirteen are publicly obtainable. Samiksha
\citep{hamna2026samiksha} is not openly redistributable; we release its schema, rubric
definitions, language coverage, evaluation prompts, and every result computed from it, and direct
interested readers to its authors, who can share it under the terms it was collected under
(App.~\ref{app:roster}).